\documentclass[11pt]{article}
\usepackage{coling2020}
\usepackage{times}
\usepackage{url}
\usepackage{latexsym}

\usepackage{adjustbox}
\usepackage{amsmath}
\usepackage{graphicx}
\usepackage{algorithm}
\usepackage{algorithmic}
\usepackage{multirow}
\usepackage{tabularx}
\usepackage{caption}
\usepackage{booktabs}
\usepackage{subcaption}
\usepackage{mwe}
\usepackage{soul}
\usepackage{url}
\usepackage{amsfonts}

\usepackage{microtype}

\usepackage{todonotes}
\makeatletter
\newcommand*\iftodonotes{\if@todonotes@disabled\expandafter\@secondoftwo\else\expandafter\@firstoftwo\fi}
\makeatother

\definecolor{edolime}{RGB}{250,218,94}

\definecolor{yaoyiranlime}{RGB}{150,200,200}

\newcolumntype{Y}{>{\centering\arraybackslash}X}

\colingfinalcopy % Uncomment this line for the final submission

\title{Emergent Communication Pretraining for Few-Shot Machine Translation}

\author{Yaoyiran Li, Edoardo M. Ponti, Ivan Vuli\'{c} and Anna Korhonen \\
  Language Technology Lab, TAL, University of Cambridge \\
  \texttt{\{yl711,ep490,iv250,alk23\}@cam.ac.uk} %\\\And
}

\date{}

\begin{document}
\maketitle
\begin{abstract}
While state-of-the-art models that rely upon massively multilingual pretrained encoders achieve sample efficiency in downstream applications, they still require abundant amounts of unlabelled text. Nevertheless, most of the world's languages lack such resources. Hence, we investigate a more radical form of unsupervised knowledge transfer in the absence of linguistic data. In particular, for the first time we pretrain neural networks via emergent communication from referential games.
%Contrary to the arbitrary nature of the lexicon,
Our key assumption is that grounding communication on images---as a crude approximation of real-world environments---inductively biases the model towards learning natural languages. 
On the one hand, we show that this substantially benefits machine translation in few-shot settings. On the other hand, this also provides an extrinsic evaluation protocol to probe the properties of emergent languages \textit{ex vitro}. Intuitively, the closer they are to natural languages, the higher the gains from pretraining on them should be. For instance, in this work we measure the influence of communication success and maximum sequence length on downstream performances. Finally, we introduce a customised adapter layer and annealing strategies for the regulariser of maximum-a-posteriori inference during fine-tuning. These turn out to be crucial to facilitate knowledge transfer and prevent catastrophic forgetting. Compared to a recurrent baseline, our method yields gains of $59.0\%$$\sim$$147.6\%$ in BLEU score with only $500$ NMT training instances and $65.1\%$$\sim$$196.7\%$ with $1,000$ NMT training instances across four language pairs. These proof-of-concept results reveal the potential of emergent communication pretraining for both natural language processing tasks in resource-poor settings and extrinsic evaluation of artificial languages.

\end{abstract}

\section{Introduction}

% MOTIVATION
Zero-shot and few-shot learning are notoriously challenging for neural networks \cite{bottou2008tradeoffs,vinyals2016matching,ravi2017optimization}. However, they are a prerequisite for natural language processing in most languages, which suffer from the paucity of annotated data \cite{ponti2019modeling}. State-of-the-art models rely on knowledge transfer, whereby an encoder is pretrained via language modeling on texts from multiple languages, and subsequently `fine-tuned' on labelled examples of resource-rich languages \cite{wu2019beto,Conneau:2020acl} or few examples in a target resource-poor language \cite{lauscher2020zero}. However, even raw texts required for pretraining are scant \cite{kornai2013digital}: for instance, Wikipedia dumps cover 278 languages out of the 7,097 spoken world-wide \cite{ethnologue-2020}.

% CONTRIBUTION 1: PRE-TRAINING WITHOUT LINGUISTIC DATA
For this reason, we push the idea of cross-lingual knowledge transfer even further, exploring and profiling a setting where not even raw natural language data for a target language are available for unsupervised pretraining. In their stead, we exploit artificial languages \textit{emerging} from a referential game on raw images \cite{kazemzadeh2014referitgame,lazaridou2016multi}. In particular, we encourage agents to cooperate in identifying images among distractors by communicating over vocabularies whose meanings are unknown. The key intuition is that, whereas lexicalisation is mostly arbitrary \cite{saussure1916cours}, communication grounded in a real-world environment does constrain what languages are likely or possible \cite{haspelmath1999optimality,croft2000explaining}. 
Hence, we hypothesise that communication over raw images offers a favourable inductive bias for natural language tasks.

% CONTRIBUTION 2: EC for NMT
In particular, we experiment with initialising an encoder-decoder model for few-shot neural machine translation with parameters pretrained on emergent communication.
In the past, emergent communication has mostly attracted theoretical interest as a tool to shed light on cooperative behaviours, the compositional properties of emergent communication protocols \cite{lazaridou2016multi,havrylov2017emergence,cao2018emergent,li2019ease,kajic2020learning}, and natural language evolution \cite{kottur2017natural,graesser2019emergent}. To our knowledge, this is the first preliminary study on deploying artificial languages from emergent communication in natural language applications. 

% CONTRIBUTION 3: NMT for EC
Conversely, our method also constitutes an extrinsic evaluation protocol to probe the properties of different emergent languages. 
The underlying assumption is that they should facilitate downstream tasks only to the extent that they share common characteristics with natural languages. In particular, we run in-depth analyses on the impact that the rate of communication success and maximum sequence length have on NMT performance.

% CONTRIBUTION 4: 
For the sake of fully leveraging the pretrained parameters and ameliorating overfitting, we also explore several new strategies to perform knowledge transfer. In particular, we customise the adapter layer \cite{pmlr-v97-houlsby19a} and propose annealing strategies for the regularisation term of MAP inference during fine-tuning. 
We run experiments in NMT between English and four languages (German, Czech, Romanian, and French) in both directions. By virtue of emergent communication pretraining and the proposed transfer strategies, we report gains in BLEU scores when simulating few-shot MT setups for the four target languages: $59.0\%$$\sim$$147.6\%$ over a standard encoder-decoder baseline when $500$ training instances are available, and $65.1\%$$\sim$$196.7\%$ when $1,000$ training instances are available. Our code is available online at \url{https://github.com/cambridgeltl/ECNMT}.
 
 %Our main contribution is threefold: (1) we are the first to adapt emergent communication pretrained models to machine translation tasks; (2) in addition to the model parameter transfer, we adopt adapter and propose $\alpha$-decay $L^{2}$-$sp$ Regularizer to further improve the results and demonstrate the effectiveness of each component; (3) we show that emergent communication pretrained model without being exposed to and human language data can efficiently improve the few-shot machine translation tasks on recurrent models and analyze the how factors such as EC prediction accuracy would influence of performance of NMT.

\begin{figure*}[t!]
\centering
\includegraphics[width=0.93\linewidth]{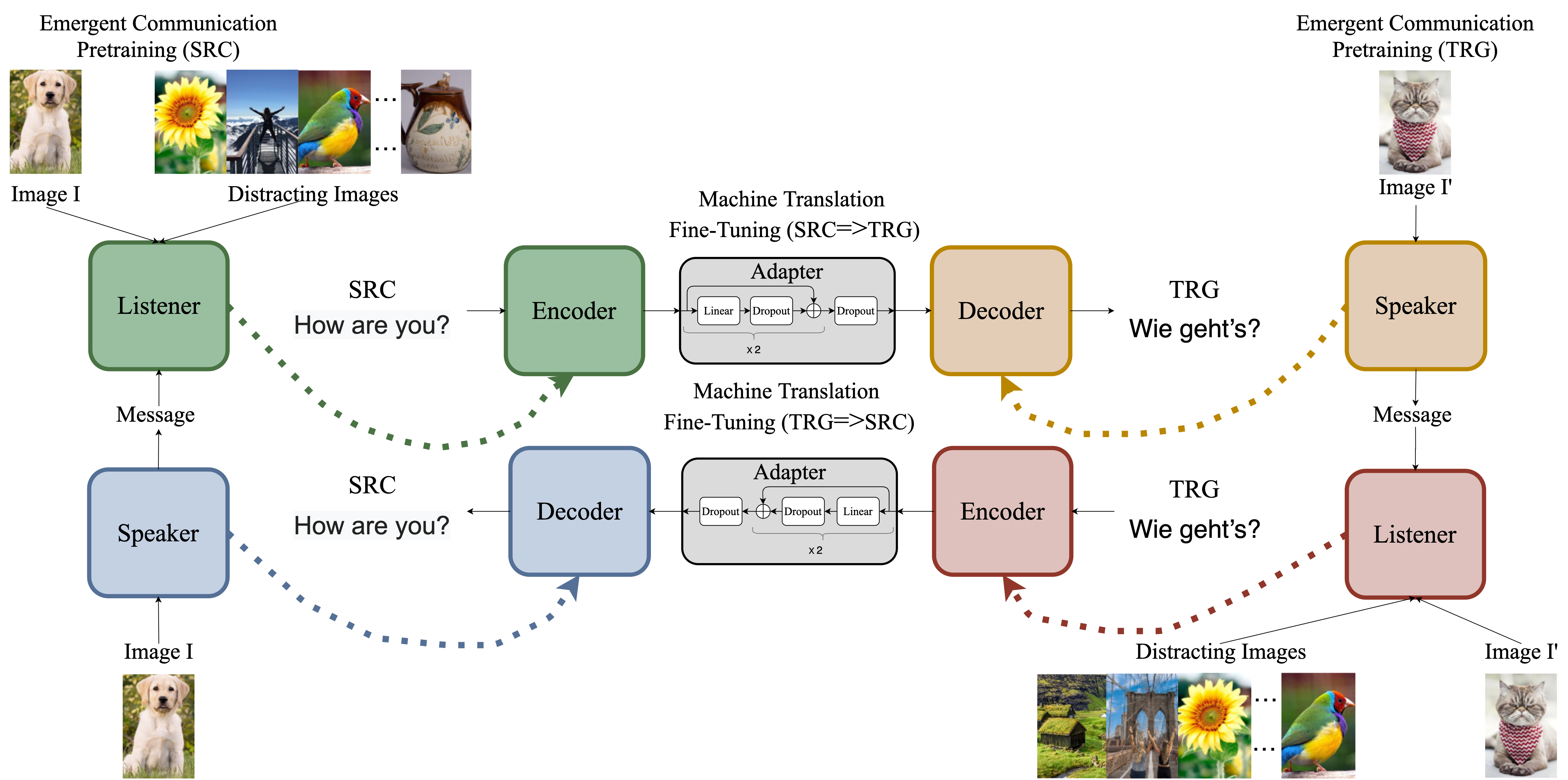}
\caption{An overview of the model architecture. Dashed lines denote parameter transfer from the EC pretraining task to the MT fine-tuning task. We stress that during EC pretraining, we do not leverage any image-caption pairs; instead, only unlabelled images are used. During MT fine-tuning, standard seq2seq NMT models are trained on SRC and TRG sentence pairs without any visual information available.}
\label{fig:arch}
\vspace{-1.5mm}
\end{figure*}

\section{Related Work}
\label{sec:rw}
Our work lies at the intersection of several prominent research areas such as pretraining for transfer learning, emergent communication, few-shot machine translation, and inductive biases for language. To all of these we cannot do full justice given space constraints.

\vspace{1.5mm}
\noindent \textbf{Pretraining for Transfer Learning.}
Unsupervised pretraining on large collections of unlabelled text yields general-purpose % static \cite{turian-etal-2010-word,Mikolov2013distributed,Bojanowski:2017tacl} or 
contextualized word representations \cite{peters2018deep,Howard:2018acl} that are beneficial across a range of downstream NLP tasks. The current dominant paradigm is training a Transformer-based deep model \cite{vaswani2017attention} relying on masked language modeling or a similar objective, as proposed in the omnipresent BERT model \cite{devlin2018bert} and its extensions \cite{Liu:2019roberta,Conneau:2019neurips,Song:2019icml,Joshi:2020spanbert}, and then fine-tuning the model further on a downstream task \cite{Wang:2019glue}. %There have been numerous extensions and improvements over the original BERT model \cite{Conneau:2019neurips,Clark:2020electra}. For instance, one line of work enriched BERT's distributional training with external knowledge from knowledge bases and linguistic resources \cite{Zhang:2019acl,liu2019kbert}. Further, MASS \cite{Song:2019icml,Song:2020arxiv}  and UniLM \cite{Dong:2019neurips} adapt BERT to generation tasks by adding auto-regressive generative objectives, while SpanBERT \cite{Joshi:2020spanbert} adapts the original BERT pretraining to span-oriented tasks. \newcite{Henderson:2019acl,Mehri:2019acl} pretrain large twin neural encoders for transfer to downstream conversational tasks.

Often this approach exploits large textual data and deep models spanning even billions of parameters \cite{Conneau:2020acl,Raffel:2019t5,Brown:2020gpt3}. In this work, we refrain from chasing task leaderboards \cite{Linzen:2020acl} and posit a fundamental question about language learning instead.
%is it possible to leverage signals from emergent communication pretraining, a setup without any linguistic data, to inform the recurrent model of a human language in a downstream task such as NMT?

\vspace{1.5mm}
\noindent \textbf{Emergent Communication.} 
The functional aspect of language \cite{clark1996using} can be captured by artificial multi-agent games \cite{Kirby:2002,Mordatch:2018aaai}, in which agents have to communicate about some shared input space (e.g., images). %There has been a recent surge of interest in communication protocols or languages that emerge from multi-agent communication oriented towards accomplishing a particular goal (e.g., detecting a correct image from an image set) \cite{foerster2016learning,li2019ease,Lowe:2019pitfalls,Singh:2019iclr,kajic2020learning}. 
A common \textit{emergent communication} protocol has been adopted in a large body of recent research: a speaker encodes a piece of information into a sequence of discrete symbols (emergent language) and a listener then aims to decipher the sequence and recover the original piece of information \cite[\textit{inter alia}]{lazaridou2016multi,havrylov2017emergence,lazaridou2018emergence,bouchacourt2018agents,chaabouni2019word,li2019ease,chaabouni2020compositionality,luna2020internal,kharitonov2020emergent}.

%%How does a language come into being and why does it evolve the way it does are old unsolved questions in human society \cite{christiansen2003language}, but it is considered that language derives meaning from its use \cite{lazaridou2018emergence,wagner2003progress,wittgenstein2009philosophical}. 

%In addition, language serves effective communication and mutual understanding between humans and language can be acquired through communication \cite{ibsen2018language}. In light of this, Emergent Communication has recently become a rising research topic in artificial multi-agent systems \cite{foerster2016learning,lazaridou2018emergence}. In recent research, a common protocol is adopted where a speaker/sender encodes a piece of information into a sequence of discrete symbols (emergent language) and a listener/receiver then aims to decipher the sequence \cite{havrylov2017emergence,lazaridou2016multi,lazaridou2018emergence,bouchacourt2018agents,chaabouni2019word,chaabouni2020compositionality,kharitonov2020emergent}. 

%% and the gradient is allowed to flow back from listener to speaking in backpropagation.
%% Another difference is that we ground machine translation in real-world understanding and thus improve the machine translation results but \cite{Lee:18} does not consider fine-tuning on translation tasks and their BLEU score drops by $10$ BLEU points on EN-DE compared with the same sequence-to-sequence baseline.

The present work is partly inspired by the work of \newcite{Lee:18}, who train agents to communicate about images with their natural language captions and use their parameters as encoder-decoders for machine translation. However, this framework relies on the availability of natural language captions (whereas we use only \textit{artificial} languages emerging from \textit{raw} images). Moreover, it does not cast EC as pretraining followed by NMT few-shot fine-tuning; rather, it learns a model in a single stage. These differences make our approach not only applicable to truly resource-lean languages but also substantially superior in performance on a same dataset such as English-German Multi30k (see \S~\ref{sec:results}). %On the other hand, \newcite{Lee:18} reported drops in performance on the same dataset. 

Another strand of recent research \cite{lowe2019learning,lowe2020interaction,lazaridou2020multi} aims at enhancing emergent communication success by encouraging agents to imitate natural language data supplied at the beginning of training. Our work goes the opposite direction and investigates whether an emergent communication protocol pretrained without any human language data can benefit downstream NLP applications such as machine translation.
  
\vspace{1.5mm}
\noindent \textbf{Few-shot Neural Machine Translation.}
Our work addresses the problem of few-shot machine translation with limited parallel data. Differently from previous methods \cite{conneau2017word,lample2017unsupervised,lample2018phrase,gu2018universal,artetxe2017unsupervised}, our approach does not draw upon auxiliary language data for pretraining, which usually consists of machine translation tasks on other languages \cite{Gu:2018emnlp} or domains \cite{sharaf2020meta}, multilingual training \cite{Aharoni:2019naacl,Liu:2020arxiv}, language model pretraining on monolingual data \cite{Conneau:2019neurips,Siddhant:2020arxiv}, back-translation techniques on monolingual data \cite{Platanios:2018emnlp,Edunov:2018emnlp}, leveraging bilingual dictionaries \cite{Duan:2020acl}, treebanks \cite{ponti-etal-2018-isomorphic}, or image captions \cite{Nakayama:2017mt,Elliott:2017ijcnlp,Lee:18}. 

On the contrary, we ground our neural model on visual knowledge acquired from agent interactions without any observation of human language, and then fine-tune our model on translation tasks even with as few as $500$ to $1,000$ training instances. We rely on few-shot MT as a standard, well-known, and sound testbed to empirically validate the crucial question of this work, that is, whether emergent communication pretraining without any natural language data can inform models of language.

\vspace{1.5mm}
\noindent \textbf{Inductive Biases for Language.}
Finally, a series of recent works has investigated how to construct neural models that are inductively biased towards learning new natural languages. This endeavour is motivated both by the need of sample efficiency and concerns of cognitive realism, as children can acquire language from limited stimuli \cite{chomsky1978naturalistic}. In particular, neural weights reflecting linguistic universals in phonotactics can be learned via approximate Bayesian inference \cite{ponti2019towards} or meta-learning \cite{mccoy2020universal}. \newcite{papadimitriou2020pretraining} found that recurrent models pretrained on non-linguistic data with latent structure (such as music or code) facilitate natural language tasks.

To our knowledge, we are the first to propose grounded communication as a non-linguistic source for pretraining, based on the hypothesis that modal and functional knowledge is a crucial inductive bias for fast and effective language acquisition.

\section{Model Architecture}
\label{s:model}
The proposed method comprises the standard two stages of transfer learning. First, as detailed in \S~\ref{ss:ec-pretraining}, we pretrain \textit{two speaker-listener agents} via emergent communication on image referential games. We then recombine\footnote{Note that we use each listener module as an MT encoder and each speaker module as a decoder. In addition, we train two separate agents because vocabulary sizes of SRC and TRG languages are different and we adopt disjoint input embeddings.} the pretrained EC agents to construct NMT encoder-decoder networks (see Figure~\ref{fig:arch}), and \textit{fine-tune} the networks on a small number of parallel sentence pairs, as we describe in \S~\ref{ssec:nmtadapt}. At the fine-tuning stage, we also add an Adapter layer between the translation encoder and decoder, and further leverage two variants of regularisation with annealing, which are outlined in \S~\ref{ssec:reganneal}. An illustrative overview of the proposed method is provided in Figure~\ref{fig:arch}. %We describe its core components in what follows.

%% (IV, this might go into footnote if there's space): 
%% when using joint dictionary or the input vocabulary sizes for source and target language are the same, it is possible to only train one)

%% (IV, said later): and adopt the loss function of \cite{Lee:18}

\subsection{Emergent Communication Pretraining}
\label{ss:ec-pretraining}
EC pretraining consists in the following referential game: an image is seen only by a \textit{speaker}, while a \textit{listener} must guess the correct image among a set of distractors based on a message generated by the speaker. Cooperation and communication therefore arise due to information asymmetry between the two players. This setup follows previous work \cite{havrylov2017emergence} with one core difference: like \newcite{Lee:18}, we train two agents, each consisting of a speaker and a listener, one for the source language, and another for the target language. Contrary to \newcite{Lee:18}, who rely on image-caption pairs for the supervised training of the speaker agents, we employ only unlabelled images to train communication protocols in an unsupervised way. The artificial language developed by agents is not explicitly constrained to resemble any natural language. We denote the two agents as ${Agent}_{s}=\{Speaker_{s},Listener_{s}\}$ and $Agent_{t}=\{Speaker_{t},Listener_{t}\}$. In our implementation, following recent work \cite{graesser2019emergent,resnick2019capacity,chaabouni2020compositionality,kharitonov2020emergent,lowe2020interaction}, both speakers and listeners are instantiated as single-layer Gated Recurrent Units (GRUs) \cite{chung2014empirical}.\footnote{We will experiment with Transformer-based architectures \cite{vaswani2017attention} in future work. Our choice of GRU is also partially motivated by recent results in few-shot MT showing on-par or even slightly stronger performance of recurrent networks over Transformers when only a small number of parallel sentences are available \cite{zhou2019handling}.} The pretraining process of $Agent_{s}$ follows these steps:

%%\footnote{Recurrent Neural Networks (RNNs) show dominance in the field of Emergent Communication \cite{graesser2019emergent,resnick2019capacity,chaabouni2020compositionality,kharitonov2020emergent,lowe2020interaction}. For recent research in few-shot machine translation, Transformers \cite{gu2018meta} and RNNs \cite{zhou2019handling}  are both common, but \cite{zhou2019handling} reports that Transformer under-performs RNN-based models in their experiments whose amount of parallel data is similar to ours. As a preliminary investigation of transferring EC pre-trained model to few-shot translation, we adopt GRU.}

\vspace{1.5mm}
\noindent \textbf{Image Set Preparation.} Let us denote the set of $N$ images as $\mathcal{D}_{I}$. At each training step, an input image $I_{i}$, $i = 1,2,...,N$ and a set of $K$ confounding images (i.e., negative examples) $C_{i} \subset  \{I_{j}|I_j\in \mathcal{D}_{I}, j \neq i\}$, $|C_{i}|=K < N$ are randomly chosen from the entire set $\mathcal{D}_{I}$. Images are represented as 2,048-dimensional feature vectors extracted from a ResNet-50 CNN \cite{He:2016cvpr}.

\vspace{1.5mm}
\noindent \textbf{Message Generation.} 
$Speaker_{s}$ takes the input image $I_{i}$ and outputs a message ${\bf m}$ describing the image, a sequence of discrete symbols of variable length. The generation comes to a halt when the special end-of-sentence symbol is emitted or the maximum message length $L_{max}$ is reached. Since ${\bf m}$ comprises discrete symbols, in order to make the model end-to-end differentiable, we adopt the Gumbel-Softmax distribution \cite{jang2016categorical,maddison2016concrete} to draw samples from categorical distributions of emergent tokens while making the gradient flow.\footnote{Another common approach is based on reinforcement learning, but recent work suggests that it is less effective and converges more slowly than Gumbel-Softmax for EC tasks \cite{havrylov2017emergence,Lee:18}.} The generation of the discrete symbol $m_t$ at each time step $t$ can be described by the following:
\begin{equation}
\begin{aligned}
{\bf h}_{t}^{s} &=
  \begin{cases}
    \text{GRU}_{\small Speaker}(\text{\textless bos\textgreater}, \text{MLP}_1(I_{i})) & t = 0 \\
    \text{GRU}_{\small Speaker}(m_{t-1},{\bf h}_{t-1}^{s}) & t > 0 \\
  \end{cases} \\
m_{t}  &= \text{Gumbel-Softmax}(\text{MLP}_2({\bf h}_{t}^{s}))
\end{aligned}
\label{formula:ec1}
\end{equation}
\noindent Here, ${\bf h}_{t}^{s}$ represents the hidden state at time step $t$, \textless bos\textgreater\ stands for the special beginning-of-sentence symbol, while MLP for multilayer perceptron. The parameters for MLP$_1$ are shared by both $Speaker$ and $Listener$ and map image features into input vectors for the GRU layer. A second MLP$_2$ is used by $Speaker$ to project each GRU hidden state---one for each time step---into vectors with dimensionality equal to the predefined vocabulary size of the emergent language.

\vspace{1.5mm}
\noindent \textbf{Image Inference.}
Given the input image, the generated message describing the image, and $K$ confounding images, $Listener_{s}$ must now guess the correct input image among the distractors. To do so, a second GRU layer decodes the message generated by the $Speaker$ as follows:
\begin{equation}
{\bf h}_{t}^{l} = 
\begin{cases}
\text{GRU}_{\small Listener}(m_{0}, {\bf 0}) & t = 0 \\
\text{GRU}_{\small Listener}(m_{t},{\bf h}_{t-1}^{l}) & t > 0 \\
\end{cases}
\label{formula:ec2}
\end{equation}
${\bf h}_{t}^{l}$ now denotes the $Listener$'s hidden state at time step $t$. The hidden state at the last time step, ${\bf h}_{|{\bf m}|}^{l}$, is used to reason over the correct image ${I}_{i}$ and $K$ distractors $C_{i}$, and guess which image is the one described by the $Speaker$. Given the message ${\bf m}$ and any image $I$ we define a compatibility score based on the inverse squared error \cite{Lee:18}:
\begin{equation}
\begin{aligned}
\text{score}({\bf m}, I) = {\left \| {\bf h}_{|{\bf m}|}^{l} - \text{MLP}_1(I) \right \|_{2}^{-2}}
%\notag
\end{aligned}
\label{formula:ec3}
\end{equation}
We then minimise the cross-entropy loss, treating the set of compatibility scores as logits, to optimise the agent parameters for the image referential game:
%
%\vspace{-2mm}
\begin{equation}
\begin{aligned}
\mathcal{L} = -\mathbb{E}_{I_{i} \in \mathcal{D}_{I}} \mathbb{E}_{\bf m} \log \underbrace{\left( \frac{e^{\text{score}({\bf m}, I_{i})}}{ \sum_{I_{j} \in \{ I_{i} \, \cup \, C_{i} \} }e^{\text{score}({\bf m}, I_{j})}} \right)}_{P(\text{guess}=I_{i}|{\bf m}, I_{i}, C_{i})}
%\notag
\end{aligned}
\label{formula:ec4}
\end{equation}
\noindent In a nutshell, $Speaker_{s}$ takes an input from the image domain, then encodes it into a message in the emergent language domain. The message conveys information that has to be transferred back to the image domain by $Listener_{s}$ in order to solve the cooperative game. The same process is repeated alternating between $Agent_{s}$ and $Agent_{t}$.\footnote{We train $Agent_{s}$ and $Agent_{t}$ separately, as in preliminary experiments we found that sharing a global $\text{MLP}_1$ for image projection does not affect NMT performance.}

\subsection{NMT Fine-Tuning and Adapters}
\label{ssec:nmtadapt}
After EC pretraining of $Agent_{s}$ and $Agent_{t}$, we recombine their speaker and listener modules into a standard sequence-to-sequence encoder-decoder neural machine translation (NMT) architecture, as shown in Figure~\ref{fig:arch}. Let us denote a training set of \textit{n} parallel sentences in the source and the target language as $\mathcal{D}$: $\{ (\mathbf{ x}^{(1)},\mathbf{y}^{(1)}), \dots, (\mathbf{ x}^{(n)},\mathbf{y}^{(n)})\}$. The model then predicts the output sequence of the $i$-th parallel sentence:

{\small
\begin{equation}
\begin{aligned}
P(\mathbf{y}^{(i)}|\mathbf{x}^{(i)}) = \prod_{t} P(y_{t}^{(i)} \mid \mathbf{y}_{<t}^{(i)},\mathbf{x}^{(i)})
\end{aligned}
\label{formula:fs01}
\end{equation}}%
\noindent In Eq.~(\ref{formula:fs01}), $\mathbf{y}_{<t}^{i}$ represents the first $t-1$ tokens in the target language sentence $\mathbf{y}^{i}$, and the input sentence $\mathbf{x}^{i}$ is encoded as a fixed-length hidden vector by the encoder, following the standard sequence-to-sequence procedure \cite{sutskever2014sequence}. The sequence loss is defined as follows:
\begin{equation}
\begin{aligned}
\mathcal{L}_{sequence} = -\mathbb{E}_{(\mathbf{ x}^{i},\mathbf{y}^{i})\in \mathcal{D}} \log P(\mathbf{y}^{i}|\mathbf{x}^{i})
\end{aligned}
\label{formula:fs02}
\end{equation}
\noindent The source-to-target translation model consists of $Listener_{s}$ (input: emergent language domain, output: image domain) and $Speaker_{t}$ (input: image domain, output: emergent language domain) and we denote their RNN parameters as $\mathbf{w^{\star}}$: these are transferred to MT fine-tuning. After fine-tuning on a small set of source-to-target sentence pairs, the model can perform the translation task. In an analogous manner, the target-to-source model is composed of $Listener_{t}$ and $Speaker_{s}$. 

To compensate for the lack of an intermediate image domain in MT, at the fine-tuning stage we add an \textit{Adapter module} in between encoders and decoders. Adapters are small neural modules that contain additional trainable parameters which facilitate quicker and more effective domain adaptation in computer vision \cite{rebuffi2017learning,rebuffi2018efficient} and, more recently, NLP tasks \cite{pmlr-v97-houlsby19a,pmlr-v97-stickland19a,pfeiffer20madx,bapna-firat-2019-simple,pfeiffer2020AdapterHub}. A notable difference compared to prior work is that during the fine-tuning stage we train jointly both the Adapter and the model parameters (which are transferred from EC). Our Adapter modules follow a simple architecture from prior work \cite{pmlr-v97-houlsby19a}, and comprise linear layers with residual connections and dropout, as illustrated in Figure~\ref{fig:arch}.

%Our adapter is simply linear layers with residual connections and dropout. Adapter is a neural module, usually residual, containing only a few trainable parameters that enables domain adaptation, which was first used in computer vision tasks \cite{rebuffi2017learning,rebuffi2018efficient}. Recent work has seen it's success in NLP tasks \cite{pmlr-v97-houlsby19a,pmlr-v97-stickland19a} where an adapter is integrated into a pre-trained BERT \cite{devlin2018bert}. We, however, apply it here in our sequence-to-sequence model and during fine-tuning stage the adapter and the transferred model parameters will be trained jointly.

%\subsection{ $\alpha$-decay $L^{2}$-$sp$ Regularizer}
\subsection{Regularisation with Annealing}
\label{ssec:reganneal}
%{\color{blue} Yaoyiran: This part is the theoretical explanation of the regularizer (different from the one shown on Thursday's meeting) that I found can improve the result significantly. Until now, I am not sure if it is theoretically correct yet. I did not find similar work in the literature. Please just consider it as a trial. If it is also problematic as the one I mentioned on Thursday's meeting, I will not hesitate to remove it and I will directly go to VI and speed up to finish VI as an important component to the model. }

%\Edo{Just to clarify my criticism last day: I am not against the idea of learning the anneal rate in itself, I was just arguing that it is something completely different from estimating the variance of the prior. So an annealing rate (learned or fixed) in principle can be \textit{combined} with whatever prior variance we choose (a hyper-parameter like $I$ or $\frac{1}{\lambda}$ or an estimate from Laplace approximation or VI).}

During fine-tuning, we also add to the objective an annealed regulariser for the encoder-decoder parameters (which, on the other hand, does not apply to the adapter module). These parameters are initialised using the parameters $\mathbf{w}^{\star}$ transferred from the EC agents. We can then define a regularisation term that prevents the parameters $\mathbf{w}$ from drifting away from their initialisation $\mathbf{w}^{\star}$ during fine-tuning \cite{duong2015low}:
\begin{equation} \label{eq:reg}
    \mathfrak{R} = \alpha \, \left \| \mathbf{w} - \mathbf{w^\star} \right \|^2
\end{equation}
\noindent where $\alpha$ is a positive real-valued tunable hyper-parameter denoting the strength of the regularisation penalty. Note that this amounts to placing a prior $\mathcal{N}(\mathbf{w}^{\star}, \mathbf{I} \alpha^{-1})$ on the encoder-decoder parameters. However, the contribution of the log-prior in Eq.~(\ref{eq:reg}) to the posterior probability of the parameters should stay fixed, whereas the contribution of the negative log-likelihood in Eq.~(\ref{formula:fs02}) should grow linearly with the number of examples. In other words, the likelihood should be able to overwhelm the prior in the limit of infinite data. 
%However, this formulation might be too stringent for longer fine-tuning. In other words, with more training steps, the model should be allowed to search parameters further away from the starting point $\mathbf{w^\star}$. 
For this reason, the importance of the regulariser should be gradually decayed over fine-tuning steps. Therefore, we propose two variants of regularisation with annealing, labelled REG-A (exponential decay) and REG-B (inverse multiplicative decay). At the fine-tuning step $k$:%\footnote{This means that when $k \to +\infty$, the regulariser gradually fades away. Since the regulariser is convex and fine-tuning stop in finite steps, we do not discuss the case of $k \to +\infty$.} 
\begin{align}
& \text{REG-A: } \mathfrak{R}(k) = \alpha \lambda^{k} \, \left \| \mathbf{w} - \mathbf{w^\star} \right \|^2 \\
%\label{formula:A}
& \text{REG-B: }    \mathfrak{R}(k) = \frac{\alpha}{k} \, \left \| \mathbf{w} - \mathbf{w^\star} \right \|^2
\label{formula:B}
\end{align}
\noindent $\lambda$ is a real-valued hyper-parameter from the interval $[0,1)$ that controls the decay steepness. The final objective function is then as follows:
\begin{equation}
\begin{aligned}
{\min_{\mathbf{w}}} \ \mathcal{L}_{sequence} + \mathfrak{R}
\end{aligned}
\end{equation}
\noindent where $\mathcal{L}_{sequence}$ is provided by Eq.~\eqref{formula:fs02}, and $\mathfrak{R}$ is one of REG-A or REG-B.

\section{Experimental Setup}

\noindent \textbf{EC Pretraining} is based on the MS COCO data set \cite{Lin:2014eccv}. We randomly select 50,000 images for training and  5,000 for validation.\footnote{Again, we stress that we do not leverage image captions (available only for few languages in COCO) at all in our setup.} For each image, a 2,048-dimensional feature vector is extracted from ResNet-50 \cite{He:2016cvpr}. The input vocabulary size for EC is equal to the BPE vocabulary size during MT fine-tuning. However, since human language data are excluded, note that there is no alignment between EC and MT BPE vocabularies.\footnote{The only exception is the end-of-sequence token $<$eos$>$.} The maximum message length, $L_{max}$, is set to an integer around the average length, in terms of BPE tokens, of the MT training sets: $15$-$18$ in our experiments. We do not impose additional constraints on the generated messages' length.\footnote{We only prevent the speakers from producing $<$eos$>$ at the beginning of their output message. Without any constraints, the messages typically occupy the entire maximum allowed length $L_{max}$.} We later profile the impact of $L_{max}$ on MT performance in \S\ref{sec:results}.

The layer size is $256$ for the input embeddings and $512$ for the hidden layers. We use Adam \cite{kingma2015adam} with a learning rate of $lr=0.001$. The dropout rate is set to $0.1$ and the Gumbel-Softmax temperature is set to $1$. The number of distracting images is $K=255$ during training, and $K=127$ in evaluation. Experiments on the validation set achieve the prediction accuracy of $>99\%$ in all EC experimental runs, i.e., the listener is able to guess the single correct image from a set of $128$ images almost always. We analyse the impact of EC prediction accuracy on few-shot MT performance later in \S\ref{sec:results}.

%% usually make full use of the max length and the average length of messages are typically in the range $(len_{max}-0.1, len_{max}]$

\vspace{1.5mm}
\noindent \textbf{Machine Translation} experiments are conducted on two standard datasets: Multi30k and Europarl. The Multi30k data set \cite{W16-3210,barrault2018findings}, originally devised for multi-modal MT, contains multilingual captions for $\approx30k$ images. We discard images and run text-only fine-tuning and evaluation on English-German (\textsc{en-de}) and English-Czech (\textsc{en-cs}) in both directions. We rely on the default training set of 29,000 pairs of parallel sentences, which we also subsample to simulate true few-shot scenarios: we randomly select 500, 1,000, and 10,000 sentence pairs for the lower-resource setups. In all experimental runs, we use the original validation set spanning 1,014 sentence pairs and the default test set spanning 1,000 pairs. 

%use EN-DE, DE-EN \cite{W16-3210}, EN-CS and CS-EN \cite{elliott-EtAl:2017:WMT} parallel sentences from Multi30k challenge task 1, where there are in total $29,000$ training sentence pairs for each language pair and we randomly choose $500$, $1,000$, $10,000$ and $29,000$ (all) sentence pairs for training.
% Europarl \cite{koehn2005epc} is crawled from proceedings of the European Parliament containing parallel text in 11 languages.

 We also run experiments on Europarl data \cite{koehn2005epc} from OPUS \cite{Tiedemann:2009opus} for two language pairs: English-Romanian (\textsc{en-ro}) and English-French (\textsc{en-fr}), again in both directions. We retain only sentences with a length between 5 and 15 words to construct data sets whose average sentence length is similar to that of Multi30k. We then randomly sample 10,000 parallel sentences as our (largest) training set, while two other disjoint random samples of 1,500 sentence pairs are used for validation and test, respectively. As with Multi30k, we again sample 500 and 1,000 training instances from the full set of 10k examples to simulate few-shot settings.

%% (IV, Removed, not needed)
%% In fact our project places particular emphasis on few-shot settings, e.g. $500$ and $1,000$ training samples, because the lack of training samples endows the method of transfer learning with more realistic function, especially for low-resource languages; in our experiments, CS and RO are relatively low-resource. Besides, we also report the case of $10,000$ training samples for better understanding our method. Additionally, we report the case of using all $29,000$ training samples for Multi30k only since itself is a small-scale date set. 
%% (IV, removed, very minor detailed, and it confuses the reader)
%% For EN-DE and DE-EN only, we adopt the BPE vocabulary used and open-sourced in previous work \cite{Lee:18} derived on the same training data, i.e. Multi30k task 1, for our own comparison. 

For each language pair, we lowercase and tokenise the data using byte-pair encoding (BPE) \cite{Sennrich2016NeuralMT}. Our BPE vocabularies are derived from all 29,000 training pairs (for the Multi30k language pairs) and 10,000 training pairs (for the Europarl language pairs). We again use Adam in the same configuration as EC pretraining, except for setting the dropout rate to $0.2$. The hyper-parameters of the annealed regulariser are set to $\alpha=5$ and $\lambda=0.998$ based on the scores on the \textsc{en-de} validation set (in the 1k training setup) and fixed to those values in all other experiments and for all other language pairs. For a fair comparison, the other hyper-parameters for fine-tuning are set identically to the NMT baseline introduced in the next paragraph.
 
 \vspace{1.5mm}
 \noindent \textbf{NMT Baseline and Evaluation Details.} The baseline NMT model is the standard seq2seq model whose architecture is exactly the same as our proposed model, but now with randomly initialised parameters (rather than transferred from EC). We extensively search the hyper-parameter space of the baseline model \cite{sennrich-zhang-2019-revisiting} and adopt Adam optimiser with learning rate of $0.001$, $\beta_1 = 0.9$, $\beta_2 = 0.999$, $\epsilon=1e\textrm{-}08$, a dropout rate of $0.2$, a batch size of $128$, a hidden-state size of $512$, an embedding size of $256$, and a max sequence length of $80$. For all models, we rely on beam search with beam size 12 for decoding. The evaluation metric is BLEU-4 \cite{Post:2018sacrebleu}.

%% (IV, removed, now added to the beginning of Section 5)
%%\paragraph{Experimental Design.} Our main experiment is to first derive EC pretrained models with prediction accuracy in the range $99.6 \pm 0.2$ on eight language pairs and then to compare our model with the baseline. In order to demonstrate the effectiveness of model parameter transfer (EC Transferred), adapter and two types of regularizers, we sequentially add adapter and regularizers on EN-DE and RO-EN to show the gain brought about by them. We also test baseline + adaper on EN-DE, DE-EN, CS-EN and RO-EN (1k samples) as an ablation study. In addition, we transfer EC pretrained models with different prediction accuracy and $len_{max}$ to investigate their influence on the gain in fine-tuning tasks.

\section{Results and Analysis}
\label{sec:results}
%Before we start our analysis, we show two findings (I did not have enough time to draw a table, but I already run a large amount of experiments to verify these): (1) generally, the higher prediction accuracy that the EC pre-training can achieve, the higher BLEU can be derived after the parameter is transferred to new task. In our EN-DE experiments our pre-trained $Agent_{s}$ achieves $99.31\%$ and $Agent_{t}$  achieves $99.6\%$ prediction accuracy; (2) the maximum length of the emergent language in the pre-training stage stage influences the machine translation performance. When the max length approximate the average length of the translation data, the performance will be satisfactory.

In what follows, we report the NMT results of our proposed model on all language pairs. We then perform an ablation study highlighting the individual contributions---of the customised adapter layer, the strategies for annealing the regulariser, and emergent communication pretraining---to the final results. Finally, we assess the impact of the rate of communication success and maximum sequence length on downstream NMT performances.

\vspace{1.5mm}
\noindent \textbf{Main Results.} The BLEU scores of the model leveraging both EC pretraining and adapters are shown in Table~\ref{table:main} for the Multi30k dataset, and in Table~\ref{table:main2} for Europarl. The results reveal sweeping gains on all language pairs and in both translation directions. These are especially accentuated in Czech and Romanian (e.g. +196.7\% in \textsc{en-cs} and +115.1\% in \textsc{ro-en} for 1k samples), which are arguably more distant from English than German and French. This suggests that our method might be particularly suited for languages that are the most challenging in real-world scenarios. Moreover, we note that the gains do not fade away as more training examples become available. For instance, while the relative improvements on the baseline decrease from +132.7\% in the 500-shot setting to +29.0\% in the 29k-shot setting (\textsc{de-en}), the absolute improvements remain consistent (+5.75 BLEU and +6.41 BLEU, respectively). Most importantly, the results clearly suggest the usefulness of EC pretraining on a downstream natural language task.

\begin{table*}[t!]
\def\arraystretch{0.99}
\begin{center}
\resizebox{1.0\textwidth}{!}{%
%{\footnotesize
\begin{tabular}{llllll}
\toprule \multicolumn{1}{c}{}  
&\multicolumn{1}{c}{\bf Model}  &\multicolumn{1}{c}{\bf 0.5k Samples}  &\multicolumn{1}{c}{\bf 1k Samples} &\multicolumn{1}{c}{\bf 10k Samples} &\multicolumn{1}{c}{\bf 29k Samples} 
\\ \hline

\multirow{3}{*}{ \rotatebox{90}{ \textsc{en-de} ~}} &\multicolumn{1}{c}{ Baseline}  &\multicolumn{1}{c}{ 4.28}  &\multicolumn{1}{c}{ 5.78} &\multicolumn{1}{c}{15.23}  &\multicolumn{1}{c}{ 20.36 } \\ 

& \multicolumn{1}{c}{EC Transferred + Adapter + REG-A}  &\multicolumn{1}{c}{   8.21}  &\multicolumn{1}{c}{  \ \ \ \ \ \ \ \ \ \ \bf 10.77  \small \small$  \uparrow^{86.3 \%}$} &\multicolumn{1}{c}{  19.93 }  &\multicolumn{1}{c}{  23.99  }\\

& \multicolumn{1}{c}{EC Transferred + Adapter + REG-B}  &\multicolumn{1}{c}{\ \ \ \ \ \ \ \ \ \ \bf 8.44 \small \small$  \uparrow^{97.1 \%}$ }  &\multicolumn{1}{c}{ 10.46} &\multicolumn{1}{c}{ \ \ \ \ \ \ \ \ \ \ \bf  21.59 \small \small$  \uparrow^{41.7 \%}$  }  &\multicolumn{1}{c}{ \ \ \ \ \ \ \ \ \ \ \bf 25.92  \small \small$  \uparrow^{27.3 \%}$}\\ \midrule

\multirow{3}{*}{ \rotatebox{90}{\textsc{de-en} ~ }} &\multicolumn{1}{c}{ Baseline}  &\multicolumn{1}{c}{ 4.33 }  &\multicolumn{1}{c}{ 6.41} &\multicolumn{1}{c}{ 15.92 }  &\multicolumn{1}{c}{ 22.09 } \\ 

& \multicolumn{1}{c}{EC Transferred + Adapter + REG-A}  &\multicolumn{1}{c}{ \ \ \ \ \ \ \ \ \ \ \bf 10.08\small \small$  \uparrow^{ 132.7\%}$}  &\multicolumn{1}{c}{  \ \ \ \ \ \ \ \ \ \ \bf  12.81 \small \small$  \uparrow^{ 99.8\%}$} &\multicolumn{1}{c}{ 20.31 }  &\multicolumn{1}{c}{ 25.65 }\\

& \multicolumn{1}{c}{EC Transferred + Adapter + REG-B}  &\multicolumn{1}{c}{ 10.04 }  &\multicolumn{1}{c}{ 12.13 } &\multicolumn{1}{c}{ \ \ \ \ \ \ \ \ \ \ \bf 22.11 \small \small$  \uparrow^{ 38.8\%}$  }  &\multicolumn{1}{c}{ \ \ \ \ \ \ \ \ \ \ \bf 28.50 \small \small$  \uparrow^{ 29.0\%}$}\\ \midrule

\multirow{3}{*}{ \rotatebox{90}{\textsc{en-cs} ~ }} &\multicolumn{1}{c}{ Baseline}  &\multicolumn{1}{c}{ 1.47 }  &\multicolumn{1}{c}{ 1.84 } &\multicolumn{1}{c}{ 9.27 }  &\multicolumn{1}{c}{ 14.73 } \\ 

& \multicolumn{1}{c}{EC Transferred + Adapter + REG-A}  &\multicolumn{1}{c}{ 3.44 }  &\multicolumn{1}{c}{  \ \ \ \ \ \ \ \ \ \ \bf 5.46\small \small$\uparrow^{ 196.7\%}$} &\multicolumn{1}{c}{ 13.33  }  &\multicolumn{1}{c}{ 17.58 }\\

& \multicolumn{1}{c}{EC Transferred + Adapter + REG-B}  &\multicolumn{1}{c}{  \ \ \ \ \ \ \ \ \ \ \bf 3.64\small \small$  \uparrow^{ 147.6\%}$}  &\multicolumn{1}{c}{ 4.96 } &\multicolumn{1}{c}{ \ \ \ \ \ \ \ \ \ \ \bf 13.62  \small \small$  \uparrow^{ 46.9\%}$  }  &\multicolumn{1}{c}{ \ \ \ \ \ \ \ \ \ \ \bf 19.07 \small \small$  \uparrow^{ 29.4\%}$}\\ \midrule

\multirow{3}{*}{ \rotatebox{90}{\textsc{cs-en} ~ }} &\multicolumn{1}{c}{ Baseline}  &\multicolumn{1}{c}{ 5.71 }  &\multicolumn{1}{c}{ 6.69 } &\multicolumn{1}{c}{ 15.15 }  &\multicolumn{1}{c}{ 19.94 } \\ 

& \multicolumn{1}{c}{EC Transferred + Adapter + REG-A}  &\multicolumn{1}{c}{\ \ \ \ \ \ \ \ \ \ \bf 9.08 \small \small$  \uparrow^{ 59.0\%}$ }  &\multicolumn{1}{c}{ 10.94 } &\multicolumn{1}{c}{ 18.56 }  &\multicolumn{1}{c}{ 22.80 }\\

& \multicolumn{1}{c}{EC Transferred + Adapter + REG-B}  &\multicolumn{1}{c}{ 8.47 }  &\multicolumn{1}{c}{\ \ \ \ \ \ \ \ \ \ \bf 11.05 \small \small $ \uparrow^{65.1\%}$ } &\multicolumn{1}{c}{\ \ \ \ \ \ \ \ \ \ \bf 19.51 \small \small$ \uparrow^{28.7\%}$  }  &\multicolumn{1}{c}{\ \ \ \ \ \ \ \ \ \ \bf 25.29 \small \small$\uparrow^{ 26.8\%}$}\\ \bottomrule
\end{tabular}
%}
}%
\vspace{-1.5mm}
\caption{BLEU scores of the full model from \S\ref{s:model} in the few-shot translation task on Multi30k with varying number of parallel sentences ($N$ Samples). $\uparrow$ represents the highest score, associated with its relative gain over the baseline.}
\label{table:main}
\end{center}
\vspace{-1.5mm}
\end{table*}
\begin{table*}[t!]
\def\arraystretch{0.99}
\begin{center}
%\resizebox{1.0\textwidth}{!}{%
{\footnotesize
\begin{tabularx}{\textwidth}{ll YYY}
\toprule \multicolumn{1}{Y}{}  
&\multicolumn{1}{c}{\bf Model}  &\multicolumn{1}{c}{\bf 0.5k Samples}  &\multicolumn{1}{c}{\bf 1k Samples} &\multicolumn{1}{c}{\bf 10k Samples} 
\\ \cmidrule(lr){2-5}

\multirow{3}{*}{ \rotatebox{90}{\textsc{en-ro} ~}} &\multicolumn{1}{c}{ Baseline}  &\multicolumn{1}{c}{ 1.71 }  &\multicolumn{1}{c}{ 2.83 } &\multicolumn{1}{c}{ 7.37 }  \\ 

& \multicolumn{1}{c}{EC Transferred + Adapter + REG-A}  &\multicolumn{1}{c}{ 3.58 }  &\multicolumn{1}{c}{\ \ \ \ \ \ \ \ \ \ \ \ \bf 5.79\small\small$\uparrow^{104.5\%}$} &\multicolumn{1}{c}{ 10.53 } \\

& \multicolumn{1}{c}{EC Transferred + Adapter + REG-B}  &\multicolumn{1}{c}{\ \ \ \ \ \ \ \ \ \ \ \ \bf 3.71\small\small$\uparrow^{116.9\%}$}  &\multicolumn{1}{c}{ 5.62 } &\multicolumn{1}{c}{\ \ \ \ \ \ \ \ \ \ \ \bf 11.35\small\small$\uparrow^{54.0\%}$} \\ \cmidrule(lr){1-5}

\multirow{3}{*}{ \rotatebox{90}{\textsc{ro-en} ~}} &\multicolumn{1}{c}{ Baseline}  &\multicolumn{1}{c}{ 1.96 }  &\multicolumn{1}{c}{ 3.03 } &\multicolumn{1}{c}{ 9.15 }  \\

& \multicolumn{1}{c}{EC Transferred + Adapter + REG-A}  &\multicolumn{1}{c}{\ \ \ \ \ \ \ \ \ \ \ \ \bf  4.49\small\small$\uparrow^{129.0\%}$}  &\multicolumn{1}{c}{ \ \ \ \ \ \ \ \ \ \ \ \ \bf 6.52\small\small$\uparrow^{115.1\%}$} &\multicolumn{1}{c}{ 12.03 } \\

& \multicolumn{1}{c}{EC Transferred + Adapter + REG-B}  &\multicolumn{1}{c}{ 4.43 }  &\multicolumn{1}{c}{ 6.10 } &\multicolumn{1}{c}{\ \ \ \ \ \ \ \ \ \ \ \ \bf 13.00 \small\small$\uparrow^{42.0\%}$} \\ \cmidrule(lr){1-5}

\multirow{3}{*}{ \rotatebox{90}{\textsc{en-fr} ~}} &\multicolumn{1}{c}{ Baseline}  &\multicolumn{1}{c}{ 1.95 }  &\multicolumn{1}{c}{ 2.50 } &\multicolumn{1}{c}{ 6.42 }  \\ 

& \multicolumn{1}{c}{EC Transferred + Adapter + REG-A}  &\multicolumn{1}{c}{ 2.96 }  &\multicolumn{1}{c}{ 4.52 } &\multicolumn{1}{c}{ 8.81 }  \\

& \multicolumn{1}{c}{EC Transferred + Adapter + REG-B}  &\multicolumn{1}{c}{\ \ \ \ \ \ \ \ \ \ \ \ \bf 3.52 \small\small$\uparrow^{80.5\%}$}  &\multicolumn{1}{c}{\ \ \ \ \ \ \ \ \ \ \ \bf 4.63\small\small$\uparrow^{85.2\%}$} &\multicolumn{1}{c}{\ \ \ \ \ \ \ \ \ \ \ \bf 9.71\small\small$\uparrow^{51.2\%}$}  \\ \cmidrule(lr){1-5}

\multirow{3}{*}{ \rotatebox{90}{\textsc{fr-en} ~}} &\multicolumn{1}{c}{ Baseline}  &\multicolumn{1}{c}{ 1.83}  &\multicolumn{1}{c}{ 2.40 } &\multicolumn{1}{c}{ 6.64 }   \\ 

& \multicolumn{1}{c}{EC Transferred + Adapter + REG-A}  &\multicolumn{1}{c}{ 3.28 }  &\multicolumn{1}{c}{ \ \ \ \ \ \ \ \ \ \ \ \bf 4.20\small\small$  \uparrow^{75.0\%}$}&\multicolumn{1}{c}{8.92}  \\

& \multicolumn{1}{c}{EC Transferred + Adapter + REG-B}  &\multicolumn{1}{c}{ \ \ \ \ \ \ \ \ \ \ \ \bf 3.64\small\small$\uparrow^{98.9\%}$}  &\multicolumn{1}{c}{ 4.12 } &\multicolumn{1}{c}{\ \ \ \ \ \ \ \ \ \ \ \bf 9.73\small\small$\uparrow^{46.5\%}$}  \\

%\multirow{2}{*}{ \rotatebox{90}{\small\ \ \ \ EN-NE}} &\multicolumn{1}{c}{ Baseline}  &\multicolumn{1}{c}{ 4.21 }  &\multicolumn{1}{c}{ 6.17 } &\multicolumn{1}{c}{ 27.07 }  &\multicolumn{1}{c}{ - } \\

%& \multicolumn{1}{c}{EC Transferred + Adapter}  &\multicolumn{1}{c}{ \bf 6.12 $\uparrow$ }  &\multicolumn{1}{c}{ \bf 8.96 $\uparrow$ } &\multicolumn{1}{c}{ \bf 29.2 $\uparrow$ }  &\multicolumn{1}{c}{ - }\\
\bottomrule
\end{tabularx}
}
\vspace{-1.5mm}
\caption{BLEU scores of the full model from \S\ref{s:model} in the few-shot translation task on Europarl with varying number of parallel sentences ($N$ Samples). $\uparrow$ represents the highest score, associated with its relative gain over the baseline.}
\label{table:main2}
\end{center}
\vspace{-1.5mm}
\end{table*}

\vspace{1.5mm}
\noindent
\textbf{Ablation Study.}
In order to disentangle the contribution of each separate component of the full model, in Table~\ref{table:ablation} we report the results on two language pairs (\textsc{en-de} and \textsc{ro-en}) for different combinations of the recurrent baseline, EC pretraining, adapters, and regulariser annealing strategies. We find that all the components improve the translation quality regardless of the amount of training data. Taking the case of \textsc{en-de} $0.5k$ as an example, the baseline achieves a BLEU of $4.28$. On top of this, EC pretraining boosts this result to $6.48$, adding the adapter layer to $5.21$. When EC and adapters are combined, they yield a BLEU of $7.52$.

Interestingly, the only finding in counter-tendency to this pattern is that the intersection of emergent communication pretraining and regulariser annealing decreases the performance compared with the baseline. Instead, when further combined with the adapters, it turns out to be the best configuration with $8.44$ BLEU. This demonstrates that EC and the adapters work in synergy and play different roles, in retaining old knowledge and in acquiring novel information, respectively.

 \begin{table*}[!t]
\def\arraystretch{0.99}
\begin{center}
%\resizebox{1.0\textwidth}{!}{%
{\footnotesize
\begin{tabularx}{1.0\textwidth}{ll XXX}
\toprule \multicolumn{1}{c}{}  
&\multicolumn{1}{c}{\bf Model}  &\multicolumn{1}{c}{\bf 0.5k Samples}  &\multicolumn{1}{c}{\bf 1k Samples} &\multicolumn{1}{c}{\bf 10k Samples} 
\\ \midrule

\multirow{8}{*}{ \rotatebox{90}{\textsc{en-de} ~}} &\multicolumn{1}{c}{ Baseline}  &\multicolumn{1}{c}{ 4.28}  &\multicolumn{1}{c}{ 5.78} &\multicolumn{1}{c}{15.23} \\ 

& \multicolumn{1}{c}{Baseline + Adapter}  &\multicolumn{1}{c}{ 5.21 }  &\multicolumn{1}{c}{ 7.25 } &\multicolumn{1}{c}{ 16.90 } \\

& \multicolumn{1}{c}{EC Transferred}  &\multicolumn{1}{c}{  6.48 }  &\multicolumn{1}{c}{ 8.47 } &\multicolumn{1}{c}{  16.33} \\

& \multicolumn{1}{c}{EC Transferred + REG-A}  &\multicolumn{1}{c}{ \ \ \  3.79 $\downarrow$}  &\multicolumn{1}{c}{\ \ \  4.88 $\downarrow$} &\multicolumn{1}{c}{16.13 }  \\

& \multicolumn{1}{c}{EC Transferred + REG-B}  &\multicolumn{1}{c}{ \ \ \  4.17 $\downarrow$}  &\multicolumn{1}{c}{\ \ \  5.72 $\downarrow$} &\multicolumn{1}{c}{  16.60 }  \\

& \multicolumn{1}{c}{EC Transferred + Adapter}  &\multicolumn{1}{c}{  7.52 }  &\multicolumn{1}{c}{ 9.25 } &\multicolumn{1}{c}{  17.59}  \\

& \multicolumn{1}{c}{EC Transferred + Adapter + REG-A}  &\multicolumn{1}{c}{   8.21}  &\multicolumn{1}{c}{ \ \ \ \ \ \ \ \ \ \ \ \bf 10.77\small\small$\uparrow^{86.3\%}$} &\multicolumn{1}{c}{  19.93 }\\

& \multicolumn{1}{c}{EC Transferred + Adapter + REG-B}  &\multicolumn{1}{c}{\ \ \ \ \ \ \ \ \ \ \ \bf8.44\small\small$\uparrow^{97.1\%}$}  &\multicolumn{1}{c}{ 10.46} &\multicolumn{1}{c}{ \ \ \ \ \ \ \ \ \ \ \ \bf  21.59\small\small$\uparrow^{41.7\%}$} \\ \midrule

\multirow{7}{*}{ \rotatebox{90}{\textsc{ro-en} ~ \ \ \ \ \ \ }} &\multicolumn{1}{c}{ Baseline}  &\multicolumn{1}{c}{ 1.96 }  &\multicolumn{1}{c}{ 3.03 } &\multicolumn{1}{c}{ 9.15 }  \\ 

& \multicolumn{1}{c}{Baseline + Adapter}  &\multicolumn{1}{c}{ 2.39 }  &\multicolumn{1}{c}{ 3.66 } &\multicolumn{1}{c}{ 9.74 } \\

& \multicolumn{1}{c}{EC Transferred}  &\multicolumn{1}{c}{  3.02 }  &\multicolumn{1}{c}{ 4.97 } &\multicolumn{1}{c}{ 10.16 }  \\

& \multicolumn{1}{c}{EC Transferred + REG-A}  &\multicolumn{1}{c}{ \ \ \  1.57 $\downarrow$}  &\multicolumn{1}{c}{\ \ \  2.12 $\downarrow$} &\multicolumn{1}{c}{\ \ \ 7.20 $\downarrow$} \\

& \multicolumn{1}{c}{EC Transferred + REG-B}  &\multicolumn{1}{c}{ \ \ \  1.09 $\downarrow$}  &\multicolumn{1}{c}{\ \ \  1.61 $\downarrow$} &\multicolumn{1}{c}{\ \ \ 8.43 $\downarrow$} \\

& \multicolumn{1}{c}{EC Transferred + Adapter}  &\multicolumn{1}{c}{\ \ \ \ \ \ \ \ \ \ \ \ \ \bf4.73\small\small$\uparrow^{~141.3\%}$}  &\multicolumn{1}{c}{ 6.11 } &\multicolumn{1}{c}{ 11.74 }  \\ 

& \multicolumn{1}{c}{EC Transferred + Adapter + REG-A}  &\multicolumn{1}{c}{   4.49}  &\multicolumn{1}{c}{\ \ \ \ \ \ \ \ \ \ \ \ \bf 6.52\small\small$\uparrow^{115.1\%}$} &\multicolumn{1}{c}{ 12.03 } \\

& \multicolumn{1}{c}{EC Transferred + Adapter + REG-B}  &\multicolumn{1}{c}{ 4.43 }  &\multicolumn{1}{c}{ 6.10 } &\multicolumn{1}{c}{\ \ \ \ \ \ \ \ \ \ \ \bf 13.00\small \small$\uparrow^{42.0\%}$}\\
\bottomrule

\end{tabularx}
%}
}
\vspace{-1.5mm}
\caption{Ablation experiments. $\downarrow$ indicates the case when an added component reduces BLEU by at least $0.4$ BLEU points; $\uparrow$ represents the highest score, associated with its relative gain over the main baseline.}
\label{table:ablation}
\end{center}
\vspace{-1.5mm}
\end{table*}

\begin{figure*}[!t]
    \centering
    \begin{subfigure}[!ht]{0.43\linewidth}
        \centering
        \includegraphics[width=0.96\linewidth]{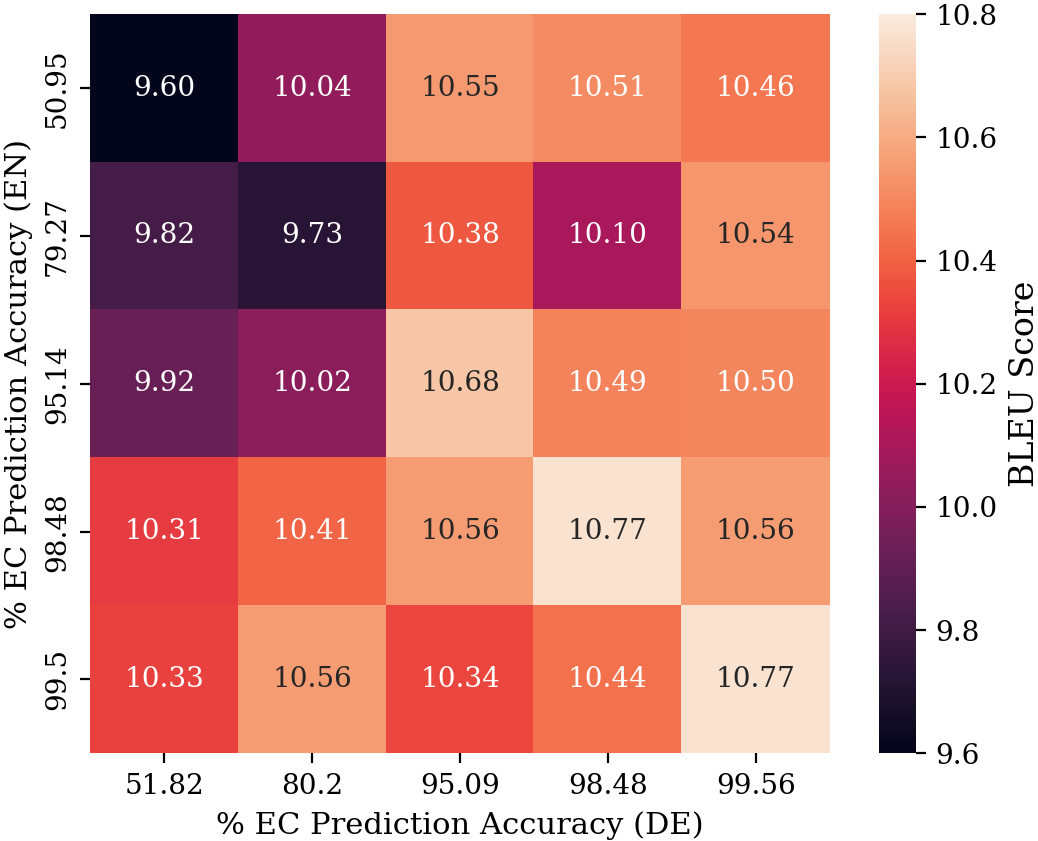}
        %\vspace{-0.7em}
        %\caption{\en--\ru: Word translation pairs}
        \label{fig:en-ru}
    \end{subfigure}
    \begin{subfigure}[!ht]{0.43\textwidth}
        \centering
        \includegraphics[width=0.96\linewidth]{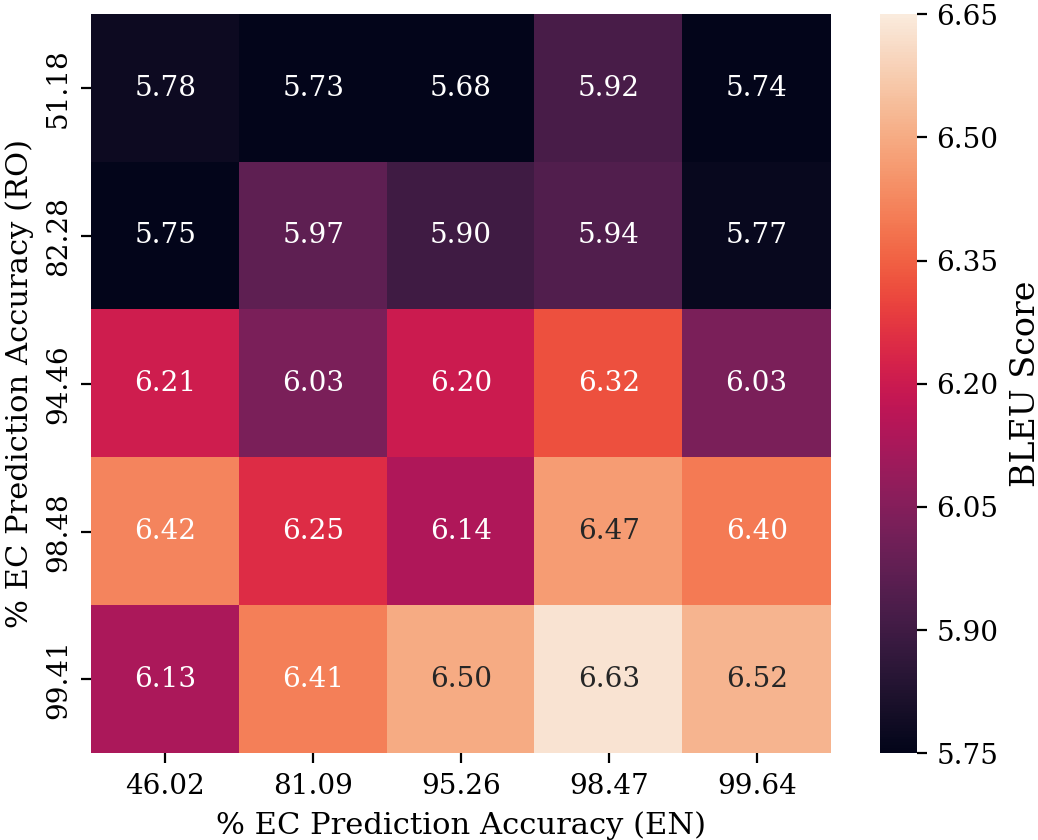}
        %\vspace{-0.7em}
        %\caption{\en--\ru: Random word pairs}
        \label{fig:en-ru-random}
    \end{subfigure}
    \vspace{-2mm}
    \caption{Impact of EC prediction accuracy on NMT BLEU scores for \textsc{en-de} (\textbf{left}) and \textsc{ro-en} (\textbf{right}). All BLEU scores are obtained in the `1k Samples' setup with the full model variant EC Transferred + Adapter + REG-A.} 
    \vspace{-1.5mm}
\label{fig:heatmap}
\label{fig:heatmap2}
\end{figure*}

\begin{figure}[t!]
\centering
\includegraphics[width=.8\textwidth]{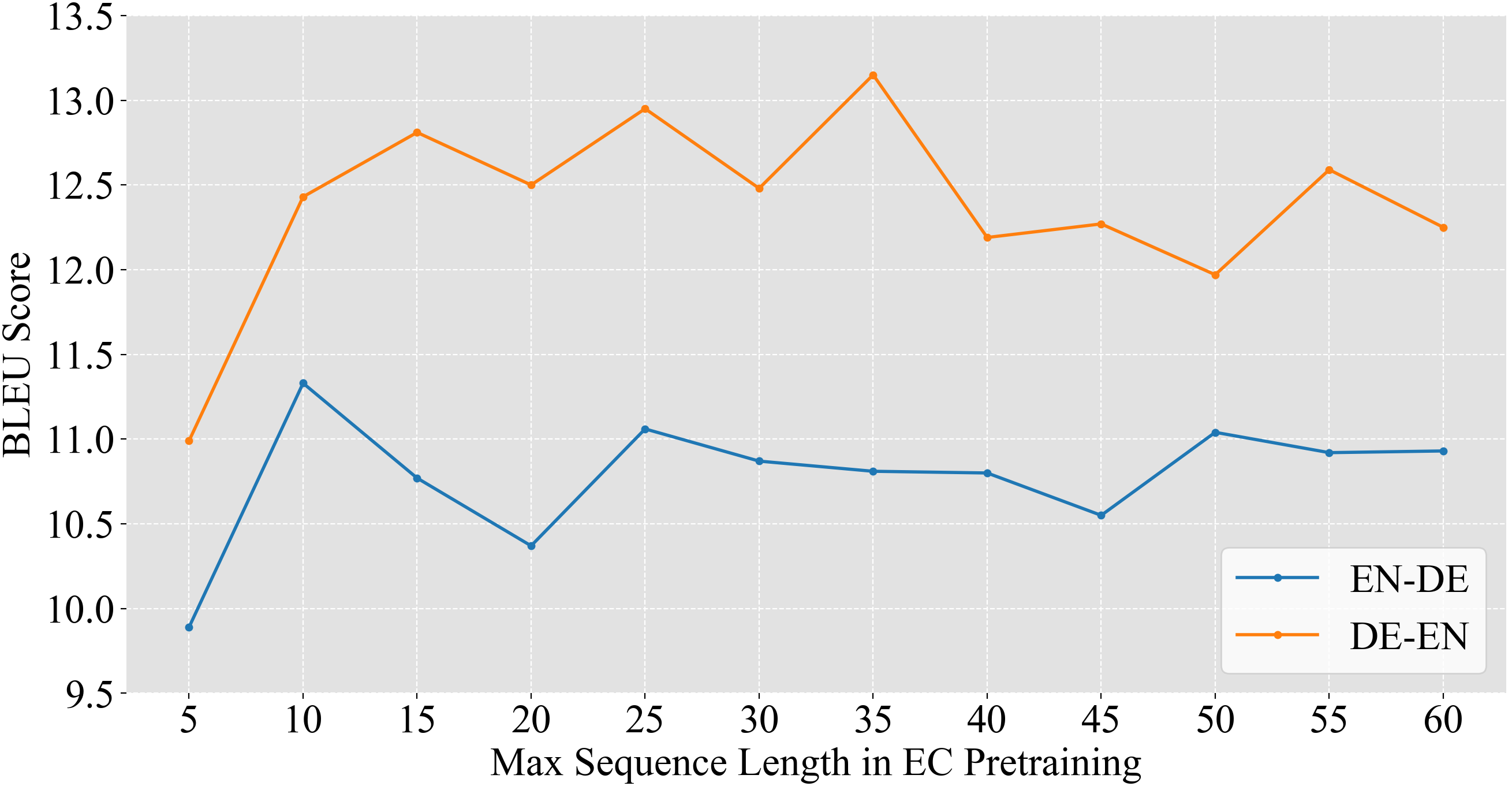}
\vspace{-1mm}
\caption{Impact of maximum EC message length ($L_{max}$) on NMT performance. All BLEU scores are obtained in the `1k Samples' setup with the full model variant EC Transferred + Adapter + REG-A.}
%%OLD-CAPTION\caption{Influence of max sequence length in EC pretraining on NMT BLEU scores. We choose max lengths ranging from $5$ to $60$, and pick up SRC and TRG EC models with prediction accuracy in $99.45 \pm 0.16$. NMT results are tested with EN-DE and DE-EN 1k samples on EC Transferred + Adapter + $\alpha$-decay A model.}
\label{fig:max_len}
\vspace{-2mm}
\end{figure}

%\begin{table}[h]
%\begin{center}
%\resizebox{0.45\textwidth}{!}{%

%\begin{tabular}{llll}
%\hline \hline  
%&\multicolumn{1}{c}{\bf Baseline+Adapter}&\multicolumn{1}{c}{\bf 1k Samples} &\multicolumn{1}{c}{\bf 10k Samples} \\ 
%\hline
%&\multicolumn{1}{c}{ EN-DE }&\multicolumn{1}{c}{ 7.25 (29.4\%) }  &\multicolumn{1}{c}{ 16.90 (26.2\%)}\\

%&\multicolumn{1}{c}{ DE-EN }&\multicolumn{1}{c}{ 7.68 (19.8\%)}  &\multicolumn{1}{c}{ 18.25 (37.6\%)}\\

%&\multicolumn{1}{c}{ CS-EN }&\multicolumn{1}{c}{ 7.31 (14.2\%) }  &\multicolumn{1}{c}{ 16.22 (24.5\%)}\\

%&\multicolumn{1}{c}{ RO-EN }&\multicolumn{1}{c}{ 3.66 (18.0\%)}  &\multicolumn{1}{c}{ 9.74 (15.3\%) }\\

%\hline
%\end{tabular}
%}
%\caption{\label{table:side} BLEU scores of Baseline + Adapter on EN-DE, DE-EN, CS-EN and RO-EN. The percentage represents the gain brought about by adapter alone on the baseline model divided by the gain resulted from EC Transfer + Adapter + decay regularizers as reported in Table \ref{table:main}.}
%\end{center}
%\end{table}

Lastly, by comparing the two strategies to anneal the regulariser during maximum-a-posteriori inference, we find no evidence favouring one or the other. Table \ref{table:main}, Table \ref{table:main2}, and Table \ref{table:ablation} show that while REG-A (exponential decay) achieves equal or better performance in 0.5k and 1k settings, REG-B (inverse multiplicative decay) shows its strength in 10k and 29k settings. A comparison on \textsc{en-de} between these two and a regulariser without annealing is shown in Appendix, where our regularisers gain an edge in all few-shot settings.\footnote{We also tried learning the prior diagonal variance via elastic weight consolidation (EWC) \cite{kirkpatrick2017overcoming} but its performance is inferior to all other regularisers in our experiments, although it remains superior to the baseline.}

%%  Therefore, we omit it from the results.

\vspace{1.5mm}
\noindent
\textbf{Influence of EC Properties on MT Fine-Tuning.}
Finally, we investigate how the properties of the artificial languages developed through EC affect downstream NMT performance. Most significantly, this can also be interpreted as a tool to evaluate whether emergent languages display affinities with natural languages. If this is the case, in fact, they should provide the correct inductive bias for NLP tasks and improve the sample efficiency of neural models.

First, we focus on the rate of communication success. During the EC pretraining stage, we save and evaluate models every $50$ training steps to pick up models with the desired level of accuracy. We run experiments on \textsc{en-de} and \textsc{ro-en} (1k samples) and select five $Agent_{s}$ and $Agent_{t}$ whose prediction accuracy is near $50\%$, $80\%$, $95\%$, $98.5\%$ and $99.5\%$, respectively. During fine-tuning we try all their $25$ possible combinations. As shown in Figure~\ref{fig:heatmap}, the prediction accuracy for $Agent_{s}$ and $Agent_{t}$ does positively correlate with MT performance. However, this trend is not strict and absolute, and sometimes sub-optimal EC models may fare better in the downstream task.

\iffalse
\begin{figure}[tbh]
\centering
\includegraphics[width=6cm]{images/heatmap.png}
\caption{Influence of EC prediction accuracy on NMT BLEU scores. We choose five SRC and TRG EC pretrained models respectively with prediction accuracies ranging from $50\%$ to $99.6\%$, NMT result tested with EN-DE 1k samples on EC Transferred + Adapter + $\alpha$-decay A model.}
\label{fig:heatmap}
\end{figure}

\begin{figure}[tbh]
\centering
\includegraphics[width=6cm]{images/roenheatmap.png}
\caption{Influence of EC prediction accuracy on NMT BLEU scores. We choose five SRC and TRG EC pretrained models respectively with prediction accuracies ranging from $46\%$ to $99.7\%$, NMT result tested with RO-EN 1k samples on EC Transferred + Adapter + $\alpha$-decay A model.}
\label{fig:heatmap2}
\end{figure}
\fi

Second, we focus on the influence of maximum sequence length. In our main results, we have set $L_{max}$ around the average sentence length (in BPE) of NMT training sets. However, we now show that this is not strictly necessary. In \textsc{en-de} and \textsc{de-en}, the average length for both languages and all splits (training, valid and test sets) are almost the same, $15$. We vary $L_{max}$  in steps of 5 from 5 to 60. In all these settings, we control for accuracy, only selecting models with a rate of communication success of $99.45\pm0.16\%$. The results are illustrated in Figure~\ref{fig:max_len}. They show that, with the exception of $L_{max} = 5$, for all the higher values MT performance is not particularly affected by $L_{max}$. %In fact, since we do not set additional conditions to constrain the length of message generated by the speaker in EC pretraining, the average message length usually approaches $len_{max}$, even when it is $60$.

%Another example is about RO-EN in our main result (Table \ref{table:main}) where we use $len_{max}$ $18$ for $Agent_{s}$ (RO) and $17$ for $Agent_{t}$ (EN); however, if we swap $18$ and $17$, then the result on $0.5k$ and $1k$ training samples will drop by circa $0.5$ BLEU point. We verify this through deriving and testing on several pretrained models. The average length of RO sentences is $18$ and EN $17$ in BPE subwords. (I need to verify this again tonight).

%(b) 300-dimensional fixed and pretrained Google Word2Vec embeddings.

%\section{A Rigorous Study of the Proposed Regularizer}
\vspace{1.5mm}
\noindent \textbf{Further Discussion.}
One interesting question concerns what kind of knowledge exactly has been learned and transferred to the fine-tuning task. Of course, the pretrained EC model does not contain any information about either SRC or TRG languages. In fact, if the adapter is trained at the MT fine-tuning stage in isolation (freezing encoder and decoder to the initialisation values), MT performance turns out to be $0$ in terms of BLEU score. What is more, it remains an open question whether the real-world grounding represented by image features plays a role in MT fine-tuning. If this were the case, one would expect higher gains in Multi30k than Europarl, as it consists of image captions. However, this does not occur in practice. As possible alternative hypotheses, EC pretraining might learn functional aspects of language \cite{wagner2003progress,wittgenstein2009philosophical,lazaridou2016multi,lazaridou2020multi}, i.e., the capability of agents to communicate and interact, or some language-universal structural properties, similar to \newcite{papadimitriou2020pretraining}. We hope that future work will shed light on this matter.

We also note that without the adapter and the regulariser, the gains on MT are relatively limited. Hence, we must additionally stress the importance and synergy of both these modules to bridge between pretraining and fine-tuning tasks. On the one hand, initialisation and regularisation avoid catastrophic forgetting of old knowledge and drifts to parameter regions unfit for communication on referential games. On the other hand, the adapter module allows a drift from the image domain and thus results in fast adaptation to the new knowledge.

\section{Conclusion and Future Work}
% Contrib 1
We have demonstrated that an extreme pretraining paradigm without any human language data, but rather based on emergent communication (EC) in referential games, provides an inductive bias for learning natural languages. In theory, it makes this paradigm applicable to any of the world's languages, most of which suffer from the paucity of annotated data. 
% Contrib 2
In particular, we focused on neural machine translation (NMT) with limited parallel data as a downstream task. Our results across several language pairs and in different few-shot setups indicate that parameter initialisations informed by EC pretraining, in combination with adapter modules and annealed regularisation, yield higher accuracy and sample efficiency than baselines without any EC pretraining.
% Contrib 3
Vice versa, we argued that NMT performance can also be interpreted as an extrinsic evaluation protocol for emergent communication models: it can assess to which extent emergent languages reflect properties found in natural languages. In particular, we discovered that communication success rate is well correlated with BLEU scores, whereas maximum sequence length is not impactful.
% Future
In the future, we plan to experiment with other state-of-the-art NMT architectures, apply our method to extremely low-resource languages, and extend the scope of our work to other tasks beyond NMT.

\section{Acknowledgements}
We thank all the anonymous reviewers for their suggestions and comments. Our work is supported by the ERC Consolidator Grant LEXICAL (no 648909). EMP, IV, and AK are also funded through the Google Faculty Research Award 2018 for Natural Language Processing.

\bibliographystyle{coling}
\bibliography{coling2020}

\begin{thebibliography}{}

\bibitem[\protect\citename{Aharoni \bgroup et al.\egroup
  }2019]{Aharoni:2019naacl}
Roee Aharoni, Melvin Johnson, and Orhan Firat.
\newblock 2019.
\newblock Massively multilingual neural machine translation.
\newblock In Jill Burstein, Christy Doran, and Thamar Solorio, editors, {\em
  Proceedings of NAACL-HLT 2019}, pages 3874--3884.

\bibitem[\protect\citename{Artetxe \bgroup et al.\egroup
  }2018]{artetxe2017unsupervised}
Mikel Artetxe, Gorka Labaka, Eneko Agirre, and Kyunghyun Cho.
\newblock 2018.
\newblock Unsupervised neural machine translation.
\newblock In {\em Proceedings of ICLR 2018}.

\bibitem[\protect\citename{Bapna and Firat}2019]{bapna-firat-2019-simple}
Ankur Bapna and Orhan Firat.
\newblock 2019.
\newblock Simple, scalable adaptation for neural machine translation.
\newblock In {\em Proceedings of EMNLP-IJCNLP 2019}, pages 1538--1548.

\bibitem[\protect\citename{Barrault \bgroup et al.\egroup
  }2018]{barrault2018findings}
Lo{\"\i}c Barrault, Fethi Bougares, Lucia Specia, Chiraag Lala, Desmond
  Elliott, and Stella Frank.
\newblock 2018.
\newblock Findings of the third shared task on multimodal machine translation.
\newblock In {\em Proceedings of the Third Conference on Machine Translation:
  Shared Task Papers}, pages 304--323.

\bibitem[\protect\citename{Bottou and Bousquet}2008]{bottou2008tradeoffs}
L{\'e}on Bottou and Olivier Bousquet.
\newblock 2008.
\newblock The tradeoffs of large scale learning.
\newblock In {\em Proceedings of NeurIPS 2008}, pages 161--168.

\bibitem[\protect\citename{Bouchacourt and Baroni}2018]{bouchacourt2018agents}
Diane Bouchacourt and Marco Baroni.
\newblock 2018.
\newblock How agents see things: {O}n visual representations in an emergent
  language game.
\newblock In {\em Proceedings of EMNLP 2018}, pages 981--985.

\bibitem[\protect\citename{Brown \bgroup et al.\egroup }2020]{Brown:2020gpt3}
Tom~B. Brown, Benjamin Mann, Nick Ryder, Melanie Subbiah, Jared Kaplan,
  Prafulla Dhariwal, Arvind Neelakantan, Pranav Shyam, Girish Sastry, Amanda
  Askell, Sandhini Agarwal, Ariel Herbert{-}Voss, Gretchen Krueger, Tom
  Henighan, Rewon Child, Aditya Ramesh, Daniel~M. Ziegler, Jeffrey Wu, Clemens
  Winter, Christopher Hesse, Mark Chen, Eric Sigler, Mateusz Litwin, Scott
  Gray, Benjamin Chess, Jack Clark, Christopher Berner, Sam McCandlish, Alec
  Radford, Ilya Sutskever, and Dario Amodei.
\newblock 2020.
\newblock Language models are few-shot learners.
\newblock In {\em Proceedings of NeurIPS 2020}.

\bibitem[\protect\citename{Cao \bgroup et al.\egroup }2018]{cao2018emergent}
Kris Cao, Angeliki Lazaridou, Marc Lanctot, Joel~Z. Leibo, Karl Tuyls, and
  Stephen Clark.
\newblock 2018.
\newblock Emergent communication through negotiation.
\newblock In {\em Proceedings of ICLR 2018}.

\bibitem[\protect\citename{Chaabouni \bgroup et al.\egroup
  }2019]{chaabouni2019word}
Rahma Chaabouni, Eugene Kharitonov, Alessandro Lazaric, Emmanuel Dupoux, and
  Marco Baroni.
\newblock 2019.
\newblock Word-order biases in deep-agent emergent communication.
\newblock In {\em Proceedings of ACL 2019}, pages 5166--5175.

\bibitem[\protect\citename{Chaabouni \bgroup et al.\egroup
  }2020]{chaabouni2020compositionality}
Rahma Chaabouni, Eugene Kharitonov, Diane Bouchacourt, Emmanuel Dupoux, and
  Marco Baroni.
\newblock 2020.
\newblock Compositionality and generalization in emergent languages.
\newblock In {\em Proceedings of ACL 2020}, pages 4427--4442.

\bibitem[\protect\citename{Chomsky}1978]{chomsky1978naturalistic}
Noam Chomsky.
\newblock 1978.
\newblock A naturalistic approach to language and cognition.
\newblock {\em Cognition and Brain Theory}, 4(1):3--22.

\bibitem[\protect\citename{Chung \bgroup et al.\egroup
  }2014]{chung2014empirical}
Junyoung Chung, Caglar Gulcehre, Kyunghyun Cho, and Yoshua Bengio.
\newblock 2014.
\newblock Empirical evaluation of gated recurrent neural networks on sequence
  modeling.
\newblock {\em arXiv preprint arXiv:1412.3555}.

\bibitem[\protect\citename{Clark}1996]{clark1996using}
Herbert~H. Clark.
\newblock 1996.
\newblock {\em Using language}.
\newblock Cambridge University Press.

\bibitem[\protect\citename{Conneau and Lample}2019]{Conneau:2019neurips}
Alexis Conneau and Guillaume Lample.
\newblock 2019.
\newblock Cross-lingual language model pretraining.
\newblock In {\em Proceedings of NeurIPS 2019}, pages 7057--7067.

\bibitem[\protect\citename{Conneau \bgroup et al.\egroup
  }2020]{Conneau:2020acl}
Alexis Conneau, Kartikay Khandelwal, Naman Goyal, Vishrav Chaudhary, Guillaume
  Wenzek, Francisco Guzm{\'{a}}n, Edouard Grave, Myle Ott, Luke Zettlemoyer,
  and Veselin Stoyanov.
\newblock 2020.
\newblock Unsupervised cross-lingual representation learning at scale.
\newblock In {\em Proceedings of ACL 2020}, pages 8440--8451.

\bibitem[\protect\citename{Croft}2000]{croft2000explaining}
William Croft.
\newblock 2000.
\newblock {\em Explaining language change: An evolutionary approach}.
\newblock Pearson Education.

\bibitem[\protect\citename{Devlin \bgroup et al.\egroup }2019]{devlin2018bert}
Jacob Devlin, Ming{-}Wei Chang, Kenton Lee, and Kristina Toutanova.
\newblock 2019.
\newblock {BERT: P}re-training of deep bidirectional transformers for language
  understanding.
\newblock In Jill Burstein, Christy Doran, and Thamar Solorio, editors, {\em
  Proceedings of NAACL-HLT 2019}, pages 4171--4186.

\bibitem[\protect\citename{Duan \bgroup et al.\egroup }2020]{Duan:2020acl}
Xiangyu Duan, Baijun Ji, Hao Jia, Min Tan, Min Zhang, Boxing Chen, Weihua Luo,
  and Yue Zhang.
\newblock 2020.
\newblock Bilingual dictionary based neural machine translation without using
  parallel sentences.
\newblock In {\em Proceedings of ACL 2020}, pages 1570--1579.

\bibitem[\protect\citename{Duong \bgroup et al.\egroup }2015]{duong2015low}
Long Duong, Trevor Cohn, Steven Bird, and Paul Cook.
\newblock 2015.
\newblock Low resource dependency parsing: Cross-lingual parameter sharing in a
  neural network parser.
\newblock In {\em Proceedings of ACL 2015}, pages 845--850.

\bibitem[\protect\citename{Eberhard \bgroup et al.\egroup
  }2020]{ethnologue-2020}
David~M. Eberhard, Gary~F. Simons, and Charles~D. Fennig, editors.
\newblock 2020.
\newblock {\em Ethnologue: Languages of the World}.
\newblock SIL International, xxiii edition.

\bibitem[\protect\citename{Edunov \bgroup et al.\egroup
  }2018]{Edunov:2018emnlp}
Sergey Edunov, Myle Ott, Michael Auli, and David Grangier.
\newblock 2018.
\newblock Understanding back-translation at scale.
\newblock In {\em Proceedings of EMNLP 2018}, pages 489--500.

\bibitem[\protect\citename{Elliott and K{\'a}d{\'a}r}2017]{Elliott:2017ijcnlp}
Desmond Elliott and {\'A}kos K{\'a}d{\'a}r.
\newblock 2017.
\newblock Imagination improves multimodal translation.
\newblock In {\em Proceedings of IJCNLP 2017}, pages 130--141.

\bibitem[\protect\citename{Elliott \bgroup et al.\egroup }2016]{W16-3210}
Desmond Elliott, Stella Frank, Khalil Sima'an, and Lucia Specia.
\newblock 2016.
\newblock Multi30k: {Multilingual English-German} image descriptions.
\newblock In {\em Proceedings of the 5th Workshop on Vision and Language},
  pages 70--74.

\bibitem[\protect\citename{Graesser \bgroup et al.\egroup
  }2019]{graesser2019emergent}
Laura Graesser, Kyunghyun Cho, and Douwe Kiela.
\newblock 2019.
\newblock Emergent linguistic phenomena in multi-agent communication games.
\newblock In Kentaro Inui, Jing Jiang, Vincent Ng, and Xiaojun Wan, editors,
  {\em Proceedings of EMNLP-IJCNLP 2019}, pages 3698--3708.

\bibitem[\protect\citename{Gu \bgroup et al.\egroup }2018a]{gu2018universal}
Jiatao Gu, Hany Hassan, Jacob Devlin, and Victor O.~K. Li.
\newblock 2018a.
\newblock Universal neural machine translation for extremely low resource
  languages.
\newblock In Marilyn~A. Walker, Heng Ji, and Amanda Stent, editors, {\em
  Proceedings of NAACL-HLT 2018}, pages 344--354.

\bibitem[\protect\citename{Gu \bgroup et al.\egroup }2018b]{Gu:2018emnlp}
Jiatao Gu, Yong Wang, Yun Chen, Victor O.~K. Li, and Kyunghyun Cho.
\newblock 2018b.
\newblock Meta-learning for low-resource neural machine translation.
\newblock In {\em Proceedings of EMNLP 2018}, pages 3622--3631.

\bibitem[\protect\citename{Haspelmath}1999]{haspelmath1999optimality}
Martin Haspelmath.
\newblock 1999.
\newblock Optimality and diachronic adaptation.
\newblock {\em Zeitschrift f{\"u}r Sprachwissenschaft}, 18(2):180--205.

\bibitem[\protect\citename{Havrylov and Titov}2017]{havrylov2017emergence}
Serhii Havrylov and Ivan Titov.
\newblock 2017.
\newblock Emergence of language with multi-agent games: Learning to communicate
  with sequences of symbols.
\newblock In {\em Proceedings of NeurIPS 2017}, pages 2149--2159.

\bibitem[\protect\citename{He \bgroup et al.\egroup }2016]{He:2016cvpr}
Kaiming He, Xiangyu Zhang, Shaoqing Ren, and Jian Sun.
\newblock 2016.
\newblock Deep residual learning for image recognition.
\newblock In {\em Proceedings of CVPR 2016}, pages 770--778.

\bibitem[\protect\citename{Houlsby \bgroup et al.\egroup
  }2019]{pmlr-v97-houlsby19a}
Neil Houlsby, Andrei Giurgiu, Stanislaw Jastrzebski, Bruna Morrone, Quentin
  De~Laroussilhe, Andrea Gesmundo, Mona Attariyan, and Sylvain Gelly.
\newblock 2019.
\newblock Parameter-efficient transfer learning for {NLP}.
\newblock In {\em Proceedings of ICML 2019}, pages 2790--2799.

\bibitem[\protect\citename{Howard and Ruder}2018]{Howard:2018acl}
Jeremy Howard and Sebastian Ruder.
\newblock 2018.
\newblock Universal language model fine-tuning for text classification.
\newblock In {\em Proceedings of ACL 2018}, pages 328--339.

\bibitem[\protect\citename{Jang \bgroup et al.\egroup
  }2017]{jang2016categorical}
Eric Jang, Shixiang Gu, and Ben Poole.
\newblock 2017.
\newblock Categorical reparameterization with {G}umbel-softmax.
\newblock In {\em Proceedings of ICLR 2017}.

\bibitem[\protect\citename{Joshi \bgroup et al.\egroup
  }2020]{Joshi:2020spanbert}
Mandar Joshi, Danqi Chen, Yinhan Liu, Daniel~S. Weld, Luke Zettlemoyer, and
  Omer Levy.
\newblock 2020.
\newblock {SpanBERT: I}mproving pre-training by representing and predicting
  spans.
\newblock {\em Transactions of the ACL}, 8:64--77.

\bibitem[\protect\citename{Kaji{\'c} \bgroup et al.\egroup
  }2020]{kajic2020learning}
Ivana Kaji{\'c}, Eser Ayg{\"u}n, and Doina Precup.
\newblock 2020.
\newblock Learning to cooperate: Emergent communication in multi-agent
  navigation.
\newblock {\em arXiv preprint arXiv:2004.01097}.

\bibitem[\protect\citename{Kazemzadeh \bgroup et al.\egroup
  }2014]{kazemzadeh2014referitgame}
Sahar Kazemzadeh, Vicente Ordonez, Mark Matten, and Tamara Berg.
\newblock 2014.
\newblock Referitgame: Referring to objects in photographs of natural scenes.
\newblock In {\em Proceedings of EMNLP 2014}, pages 787--798.

\bibitem[\protect\citename{Kharitonov and Baroni}2020]{kharitonov2020emergent}
Eugene Kharitonov and Marco Baroni.
\newblock 2020.
\newblock Emergent language generalization and acquisition speed are not tied
  to compositionality.
\newblock {\em arXiv preprint arXiv:2004.03420}.

\bibitem[\protect\citename{Kingma and Ba}2015]{kingma2015adam}
Diederik~P Kingma and Jimmy Ba.
\newblock 2015.
\newblock Adam: A method for stochastic optimization.
\newblock In {\em Proceedings of ICLR 2015}.

\bibitem[\protect\citename{Kirby}2002]{Kirby:2002}
Simon Kirby.
\newblock 2002.
\newblock Natural language from artificial life.
\newblock {\em Artif. Life}, 8(2):185--215.

\bibitem[\protect\citename{Kirkpatrick \bgroup et al.\egroup
  }2017]{kirkpatrick2017overcoming}
James Kirkpatrick, Razvan Pascanu, Neil Rabinowitz, Joel Veness, Guillaume
  Desjardins, Andrei~A Rusu, Kieran Milan, John Quan, Tiago Ramalho, Agnieszka
  Grabska-Barwinska, et~al.
\newblock 2017.
\newblock Overcoming catastrophic forgetting in neural networks.
\newblock {\em Proceedings of the National Academy of Sciences},
  114(13):3521--3526.

\bibitem[\protect\citename{Koehn}2005]{koehn2005epc}
Philipp Koehn.
\newblock 2005.
\newblock Europarl: {A} parallel corpus for statistical machine translation.
\newblock In {\em Proceedings of the 10th Machine Translation Summit}, pages
  79--86.

\bibitem[\protect\citename{Kornai}2013]{kornai2013digital}
Andr{\'a}s Kornai.
\newblock 2013.
\newblock Digital language death.
\newblock {\em PLoS One}, 8(10):e77056.

\bibitem[\protect\citename{Kottur \bgroup et al.\egroup
  }2017]{kottur2017natural}
Satwik Kottur, Jos{\'{e}} M.~F. Moura, Stefan Lee, and Dhruv Batra.
\newblock 2017.
\newblock Natural language does not emerge 'naturally' in multi-agent dialog.
\newblock In {\em Proceedings of EMNLP 2017}, pages 2962--2967.

\bibitem[\protect\citename{Lample \bgroup et al.\egroup
  }2018a]{lample2017unsupervised}
Guillaume Lample, Alexis Conneau, Ludovic Denoyer, and Marc'Aurelio Ranzato.
\newblock 2018a.
\newblock Unsupervised machine translation using monolingual corpora only.
\newblock In {\em Proceedings of ICLR 2018}.

\bibitem[\protect\citename{Lample \bgroup et al.\egroup
  }2018b]{conneau2017word}
Guillaume Lample, Alexis Conneau, Marc'Aurelio Ranzato, Ludovic Denoyer, and
  Herv{\'{e}} J{\'{e}}gou.
\newblock 2018b.
\newblock Word translation without parallel data.
\newblock In {\em Proceedings of ICLR 2018}.

\bibitem[\protect\citename{Lample \bgroup et al.\egroup
  }2018c]{lample2018phrase}
Guillaume Lample, Myle Ott, Alexis Conneau, Ludovic Denoyer, and Marc'Aurelio
  Ranzato.
\newblock 2018c.
\newblock Phrase-based {\&} neural unsupervised machine translation.
\newblock In {\em Proceedings of EMNLP 2018}, pages 5039--5049.

\bibitem[\protect\citename{Lauscher \bgroup et al.\egroup
  }2020]{lauscher2020zero}
Anne Lauscher, Vinit Ravishankar, Ivan Vuli{\'c}, and Goran Glava{\v{s}}.
\newblock 2020.
\newblock From zero to hero: {O}n the limitations of zero-shot cross-lingual
  transfer with multilingual transformers.
\newblock In {\em Proceedings of EMNLP 2020}.

\bibitem[\protect\citename{Lazaridou \bgroup et al.\egroup
  }2017]{lazaridou2016multi}
Angeliki Lazaridou, Alexander Peysakhovich, and Marco Baroni.
\newblock 2017.
\newblock Multi-agent cooperation and the emergence of (natural) language.
\newblock In {\em Proceedings of ICLR 2017}.

\bibitem[\protect\citename{Lazaridou \bgroup et al.\egroup
  }2018]{lazaridou2018emergence}
Angeliki Lazaridou, Karl~Moritz Hermann, Karl Tuyls, and Stephen Clark.
\newblock 2018.
\newblock Emergence of linguistic communication from referential games with
  symbolic and pixel input.
\newblock In {\em Proceedings of ICLR 2018}.

\bibitem[\protect\citename{Lazaridou \bgroup et al.\egroup
  }2020]{lazaridou2020multi}
Angeliki Lazaridou, Anna Potapenko, and Olivier Tieleman.
\newblock 2020.
\newblock Multi-agent communication meets natural language: Synergies between
  functional and structural language learning.
\newblock In {\em Proceedings of ACL 2020}, pages 7663--7674.

\bibitem[\protect\citename{Lee \bgroup et al.\egroup }2018]{Lee:18}
Jason Lee, Kyunghyun Cho, Jason Weston, and Douwe Kiela.
\newblock 2018.
\newblock Emergent translation in multi-agent communication.
\newblock In {\em Proceedings of ICLR 2018}.

\bibitem[\protect\citename{Li and Bowling}2019]{li2019ease}
Fushan Li and Michael Bowling.
\newblock 2019.
\newblock Ease-of-teaching and language structure from emergent communication.
\newblock In {\em Proceedings of NeurIPS 2019}, pages 15825--15835.

\bibitem[\protect\citename{Lin \bgroup et al.\egroup }2014]{Lin:2014eccv}
Tsung{-}Yi Lin, Michael Maire, Serge~J. Belongie, James Hays, Pietro Perona,
  Deva Ramanan, Piotr Doll{\'{a}}r, and C.~Lawrence Zitnick.
\newblock 2014.
\newblock Microsoft {COCO: C}ommon objects in context.
\newblock In {\em Proceedings of ECCV 2014}, pages 740--755.

\bibitem[\protect\citename{Linzen}2020]{Linzen:2020acl}
Tal Linzen.
\newblock 2020.
\newblock How can we accelerate progress towards human-like linguistic
  generalization?
\newblock In {\em Proceedings of ACL 2020}, pages 5210--5217.

\bibitem[\protect\citename{Liu \bgroup et al.\egroup }2019]{Liu:2019roberta}
Yinhan Liu, Myle Ott, Naman Goyal, Jingfei Du, Mandar Joshi, Danqi Chen, Omer
  Levy, Mike Lewis, Luke Zettlemoyer, and Veselin Stoyanov.
\newblock 2019.
\newblock {RoBERTa:} {A} robustly optimized {BERT} pretraining approach.
\newblock {\em CoRR}, abs/1907.11692.

\bibitem[\protect\citename{Liu \bgroup et al.\egroup }2020]{Liu:2020arxiv}
Yinhan Liu, Jiatao Gu, Naman Goyal, Xian Li, Sergey Edunov, Marjan
  Ghazvininejad, Mike Lewis, and Luke Zettlemoyer.
\newblock 2020.
\newblock Multilingual denoising pre-training for neural machine translation.
\newblock {\em CoRR}, abs/2001.08210.

\bibitem[\protect\citename{Lowe \bgroup et al.\egroup }2019]{lowe2019learning}
Ryan Lowe, Abhinav Gupta, Jakob Foerster, Douwe Kiela, and Joelle Pineau.
\newblock 2019.
\newblock Learning to learn to communicate.
\newblock In {\em Proceedings of the 1st Adaptive \& Multitask Learning
  Workshop}.

\bibitem[\protect\citename{Lowe \bgroup et al.\egroup
  }2020]{lowe2020interaction}
Ryan Lowe, Abhinav Gupta, Jakob Foerster, Douwe Kiela, and Joelle Pineau.
\newblock 2020.
\newblock On the interaction between supervision and self-play in emergent
  communication.
\newblock In {\em Proceedings of ICLR 2020}.

\bibitem[\protect\citename{Luna \bgroup et al.\egroup }2020]{luna2020internal}
Diana~Rodr{\'\i}guez Luna, Edoardo~Maria Ponti, Dieuwke Hupkes, and Elia Bruni.
\newblock 2020.
\newblock Internal and external pressures on language emergence: {L}east
  effort, object constancy and frequency.
\newblock {\em arXiv preprint arXiv:2004.03868}.

\bibitem[\protect\citename{Maddison \bgroup et al.\egroup
  }2017]{maddison2016concrete}
Chris~J. Maddison, Andriy Mnih, and Yee~Whye Teh.
\newblock 2017.
\newblock The concrete distribution: {A} continuous relaxation of discrete
  random variables.
\newblock In {\em Proceedings of ICLR 2017}.

\bibitem[\protect\citename{McCoy \bgroup et al.\egroup
  }2020]{mccoy2020universal}
R.~Thomas McCoy, Erin Grant, Paul Smolensky, Thomas~L Griffiths, and Tal
  Linzen.
\newblock 2020.
\newblock Universal linguistic inductive biases via meta-learning.
\newblock In {\em Proceedings of CogSci 2020}.

\bibitem[\protect\citename{Mordatch and Abbeel}2018]{Mordatch:2018aaai}
Igor Mordatch and Pieter Abbeel.
\newblock 2018.
\newblock Emergence of grounded compositional language in multi-agent
  populations.
\newblock In {\em Proceedings of AAAI}, pages 1495--1502.

\bibitem[\protect\citename{Nakayama and Nishida}2017]{Nakayama:2017mt}
Hideki Nakayama and Noriki Nishida.
\newblock 2017.
\newblock Zero-resource machine translation by multimodal encoder-decoder
  network with multimedia pivot.
\newblock {\em Machine Translation}, 31(1-2):49--64.

\bibitem[\protect\citename{Papadimitriou and
  Jurafsky}2020]{papadimitriou2020pretraining}
Isabel Papadimitriou and Dan Jurafsky.
\newblock 2020.
\newblock Pretraining on non-linguistic structure as a tool for analyzing
  learning bias in language models.
\newblock {\em arXiv preprint arXiv:2004.14601}.

\bibitem[\protect\citename{Peters \bgroup et al.\egroup }2018]{peters2018deep}
Matthew~E. Peters, Mark Neumann, Mohit Iyyer, Matt Gardner, Christopher Clark,
  Kenton Lee, and Luke Zettlemoyer.
\newblock 2018.
\newblock Deep contextualized word representations.
\newblock In {\em Proceedings of NAACL-HLT 2018}, pages 2227--2237.

\bibitem[\protect\citename{Pfeiffer \bgroup et al.\egroup
  }2020a]{pfeiffer2020AdapterHub}
Jonas Pfeiffer, Andreas R\"uckl\'{e}, Clifton Poth, Aishwarya Kamath, Ivan
  Vuli\'{c}, Sebastian Ruder, Kyunghyun Cho, and Iryna Gurevych.
\newblock 2020a.
\newblock {AdapterHub: A} framework for adapting {T}ransformers.
\newblock In {\em Proceedings of EMNLP 2020: System Demonstrations}.

\bibitem[\protect\citename{Pfeiffer \bgroup et al.\egroup
  }2020b]{pfeiffer20madx}
Jonas Pfeiffer, Ivan Vuli\'{c}, Iryna Gurevych, and Sebastian Ruder.
\newblock 2020b.
\newblock {MAD-X: A}n adapter-based framework for multi-task cross-lingual
  transfer.
\newblock In {\em Proceedings of EMNLP 2020}.

\bibitem[\protect\citename{Platanios \bgroup et al.\egroup
  }2018]{Platanios:2018emnlp}
Emmanouil~Antonios Platanios, Mrinmaya Sachan, Graham Neubig, and Tom~M.
  Mitchell.
\newblock 2018.
\newblock Contextual parameter generation for universal neural machine
  translation.
\newblock In {\em Proceedings of EMNLP 2018}, pages 425--435.

\bibitem[\protect\citename{Ponti \bgroup et al.\egroup
  }2018]{ponti-etal-2018-isomorphic}
Edoardo~Maria Ponti, Roi Reichart, Anna Korhonen, and Ivan Vuli{\'c}.
\newblock 2018.
\newblock Isomorphic transfer of syntactic structures in cross-lingual {NLP}.
\newblock In {\em Proceedings of ACL 2018}, pages 1531--1542.

\bibitem[\protect\citename{Ponti \bgroup et al.\egroup
  }2019a]{ponti2019modeling}
Edoardo~Maria Ponti, Helen O'Horan, Yevgeni Berzak, Ivan Vuli{\'c}, Roi
  Reichart, Thierry Poibeau, Ekaterina Shutova, and Anna Korhonen.
\newblock 2019a.
\newblock Modeling language variation and universals: A survey on typological
  linguistics for natural language processing.
\newblock {\em Computational Linguistics}, 45(3):559--601.

\bibitem[\protect\citename{Ponti \bgroup et al.\egroup
  }2019b]{ponti2019towards}
Edoardo~Maria Ponti, Ivan Vuli{\'c}, Ryan Cotterell, Roi Reichart, and Anna
  Korhonen.
\newblock 2019b.
\newblock Towards zero-shot language modeling.
\newblock In {\em Proceedings of EMNLP-IJCNLP 2019}, pages 2893--2903.

\bibitem[\protect\citename{Post}2018]{Post:2018sacrebleu}
Matt Post.
\newblock 2018.
\newblock A call for clarity in reporting {BLEU} scores.
\newblock In {\em Proceedings of the Third Conference on Machine Translation:
  Research Papers}, pages 186--191.

\bibitem[\protect\citename{Raffel \bgroup et al.\egroup }2019]{Raffel:2019t5}
Colin Raffel, Noam Shazeer, Adam Roberts, Katherine Lee, Sharan Narang, Michael
  Matena, Yanqi Zhou, Wei Li, and Peter~J. Liu.
\newblock 2019.
\newblock Exploring the limits of transfer learning with a unified text-to-text
  transformer.
\newblock {\em CoRR}, abs/1910.10683.

\bibitem[\protect\citename{Ravi and Larochelle}2017]{ravi2017optimization}
Sachin Ravi and Hugo Larochelle.
\newblock 2017.
\newblock Optimization as a model for few-shot learning.
\newblock In {\em Proceedings of ICLR 2017}.

\bibitem[\protect\citename{Rebuffi \bgroup et al.\egroup
  }2017]{rebuffi2017learning}
Sylvestre-Alvise Rebuffi, Hakan Bilen, and Andrea Vedaldi.
\newblock 2017.
\newblock Learning multiple visual domains with residual adapters.
\newblock In {\em Proceedings of NeurIPS 2017}, pages 506--516.

\bibitem[\protect\citename{Rebuffi \bgroup et al.\egroup
  }2018]{rebuffi2018efficient}
Sylvestre-Alvise Rebuffi, Hakan Bilen, and Andrea Vedaldi.
\newblock 2018.
\newblock Efficient parametrization of multi-domain deep neural networks.
\newblock In {\em Proceedings of CVPR 2018}, pages 8119--8127.

\bibitem[\protect\citename{Resnick \bgroup et al.\egroup
  }2020]{resnick2019capacity}
Cinjon Resnick, Abhinav Gupta, Jakob~N. Foerster, Andrew~M. Dai, and Kyunghyun
  Cho.
\newblock 2020.
\newblock Capacity, bandwidth, and compositionality in emergent language
  learning.
\newblock In {\em Proceedings of AAMAS 2020}, pages 1125--1133.

\bibitem[\protect\citename{Saussure}1916]{saussure1916cours}
Ferdinand~de Saussure.
\newblock 1916.
\newblock {\em Cours de linguistique g{\'e}n{\'e}rale, ed}.
\newblock Payot.
\newblock Edited by C. Bally and A. Sechehaye.

\bibitem[\protect\citename{Sennrich and
  Zhang}2019]{sennrich-zhang-2019-revisiting}
Rico Sennrich and Biao Zhang.
\newblock 2019.
\newblock Revisiting low-resource neural machine translation: A case study.
\newblock In {\em Proceedings of ACL 2019}, pages 211--221.

\bibitem[\protect\citename{Sennrich \bgroup et al.\egroup
  }2016]{Sennrich2016NeuralMT}
Rico Sennrich, Barry Haddow, and Alexandra Birch.
\newblock 2016.
\newblock Neural machine translation of rare words with subword units.
\newblock In {\em Proceedings of ACL 2016}, pages 1715--1725.

\bibitem[\protect\citename{Sharaf \bgroup et al.\egroup }2020]{sharaf2020meta}
Amr Sharaf, Hany Hassan, and Hal Daum{\'e}~III.
\newblock 2020.
\newblock Meta-learning for few-shot {NMT} adaptation.
\newblock {\em arXiv preprint arXiv:2004.02745}.

\bibitem[\protect\citename{Siddhant \bgroup et al.\egroup
  }2020]{Siddhant:2020arxiv}
Aditya Siddhant, Ankur Bapna, Yuan Cao, Orhan Firat, Mia~Xu Chen, Sneha~Reddy
  Kudugunta, Naveen Arivazhagan, and Yonghui Wu.
\newblock 2020.
\newblock Leveraging monolingual data with self-supervision for multilingual
  neural machine translation.
\newblock {\em CoRR}, abs/2005.04816.

\bibitem[\protect\citename{Song \bgroup et al.\egroup }2019]{Song:2019icml}
Kaitao Song, Xu~Tan, Tao Qin, Jianfeng Lu, and Tie{-}Yan Liu.
\newblock 2019.
\newblock {MASS: M}asked sequence to sequence pre-training for language
  generation.
\newblock In {\em Proceedings of ICML 2019}, volume~97, pages 5926--5936.

\bibitem[\protect\citename{Stickland and Murray}2019]{pmlr-v97-stickland19a}
Asa~Cooper Stickland and Iain Murray.
\newblock 2019.
\newblock {BERT} and {PAL}s: Projected attention layers for efficient
  adaptation in multi-task learning.
\newblock In {\em Proceedings of ICML 2019}, pages 5986--5995.

\bibitem[\protect\citename{Sutskever \bgroup et al.\egroup
  }2014]{sutskever2014sequence}
Ilya Sutskever, Oriol Vinyals, and Quoc~V Le.
\newblock 2014.
\newblock Sequence to sequence learning with neural networks.
\newblock In {\em Proceedings of NeurIPS 2014}, pages 3104--3112.

\bibitem[\protect\citename{Tiedemann}2009]{Tiedemann:2009opus}
J\"org Tiedemann.
\newblock 2009.
\newblock News from {OPUS} - {A} collection of multilingual parallel corpora
  with tools and interfaces.
\newblock In {\em Proceedings of RANLP 2009}, pages 237--248.

\bibitem[\protect\citename{Vaswani \bgroup et al.\egroup
  }2017]{vaswani2017attention}
Ashish Vaswani, Noam Shazeer, Niki Parmar, Jakob Uszkoreit, Llion Jones,
  Aidan~N Gomez, {\L}ukasz Kaiser, and Illia Polosukhin.
\newblock 2017.
\newblock Attention is all you need.
\newblock In {\em Proceedings of NeurIPS 2017}, pages 5998--6008.

\bibitem[\protect\citename{Vinyals \bgroup et al.\egroup
  }2016]{vinyals2016matching}
Oriol Vinyals, Charles Blundell, Timothy Lillicrap, Daan Wierstra, et~al.
\newblock 2016.
\newblock Matching networks for one shot learning.
\newblock In {\em Proceedings of NeurIPS 2016}, pages 3630--3638.

\bibitem[\protect\citename{Wagner \bgroup et al.\egroup
  }2003]{wagner2003progress}
Kyle Wagner, James~A Reggia, Juan Uriagereka, and Gerald~S Wilkinson.
\newblock 2003.
\newblock Progress in the simulation of emergent communication and language.
\newblock {\em Adaptive Behavior}, 11(1):37--69.

\bibitem[\protect\citename{Wang \bgroup et al.\egroup }2019]{Wang:2019glue}
Alex Wang, Amanpreet Singh, Julian Michael, Felix Hill, Omer Levy, and
  Samuel~R. Bowman.
\newblock 2019.
\newblock {GLUE:} {A} multi-task benchmark and analysis platform for natural
  language understanding.
\newblock In {\em Proceedings of ICLR 2019}.

\bibitem[\protect\citename{Wittgenstein}2009]{wittgenstein2009philosophical}
Ludwig Wittgenstein.
\newblock 2009.
\newblock {\em Philosophical investigations}.
\newblock John Wiley \& Sons.

\bibitem[\protect\citename{Wu and Dredze}2019]{wu2019beto}
Shijie Wu and Mark Dredze.
\newblock 2019.
\newblock Beto, bentz, becas: The surprising cross-lingual effectiveness of
  bert.
\newblock In {\em Proceedings of EMNLP-IJCNLP 2019}, pages 833--844.

\bibitem[\protect\citename{Zhou \bgroup et al.\egroup }2019]{zhou2019handling}
Chunting Zhou, Xuezhe Ma, Junjie Hu, and Graham Neubig.
\newblock 2019.
\newblock Handling syntactic divergence in low-resource machine translation.
\newblock In {\em Proceedings of EMNLP-IJCNLP 2019}, pages 1388--1394.

\end{thebibliography}

\clearpage
\appendix

\section{Appendices}
\label{sec:appendix}

\subsection{Regulariser without Annealing}

In Table~\ref{table:appendix}, we compare the BLEU scores of our proposed annealed regularisers (REG-A and REG-B) and a regulariser without annealing (NA) on \textsc{en-de}. $\downarrow$ indicates when the newly added module reduces the BLEU score by at least $0.4$ BLEU points, and $\uparrow$ represents the highest gain compared with the baseline.

\begin{table}[h]
\begin{center}
\resizebox{1.0\textwidth}{!}{%
\begin{tabular}{llllll}
\toprule \multicolumn{1}{c}{}  
&\multicolumn{1}{c}{\bf Model}  &\multicolumn{1}{c}{\bf 0.5k Samples}  &\multicolumn{1}{c}{\bf 1k Samples} &\multicolumn{1}{c}{\bf 10k Samples} &\multicolumn{1}{c}{\bf 29k Samples} 
\\ \hline

\multirow{17}{*}{ \rotatebox{90}{\small EN-DE ~}} &\multicolumn{1}{c}{ Baseline}  &\multicolumn{1}{c}{ 4.28}  &\multicolumn{1}{c}{ 5.78} &\multicolumn{1}{c}{15.23}  &\multicolumn{1}{c}{ 20.36 } \\

& \multicolumn{1}{c}{EC Transferred}  &\multicolumn{1}{c}{  6.48 }  &\multicolumn{1}{c}{ 8.47 } &\multicolumn{1}{c}{  16.33}  &\multicolumn{1}{c}{  21.43 }\\

& \multicolumn{1}{c}{EC Transferred + NA, $\alpha$ = 5e-3}  &\multicolumn{1}{c}{\ \ \  3.51 $\downarrow$}  &\multicolumn{1}{c}{\ \ \  5.01 $\downarrow$} &\multicolumn{1}{c}{\ \ \  7.36 $\downarrow$ }  &\multicolumn{1}{c}{\ \ \  8.07 $\downarrow$ }\\

& \multicolumn{1}{c}{EC Transferred + NA, $\alpha$ = 5e-4}  &\multicolumn{1}{c}{\ \ \  5.42 $\downarrow$}  &\multicolumn{1}{c}{\ \ \  6.94 $\downarrow$} &\multicolumn{1}{c}{16.69}  &\multicolumn{1}{c}{21.84}\\

& \multicolumn{1}{c}{EC Transferred + NA, $\alpha$ = 5e-5}  &\multicolumn{1}{c}{6.31}  &\multicolumn{1}{c}{\ \ \  7.98 $\downarrow$ } &\multicolumn{1}{c}{16.36}  &\multicolumn{1}{c}{22.93}\\

& \multicolumn{1}{c}{EC Transferred + NA, $\alpha$ = 5e-6}  &\multicolumn{1}{c}{6.8}  &\multicolumn{1}{c}{9.35} &\multicolumn{1}{c}{16.35}  &\multicolumn{1}{c}{21.89}\\

& \multicolumn{1}{c}{EC Transferred + NA, $\alpha$ = 5e-7}  &\multicolumn{1}{c}{7.02}  &\multicolumn{1}{c}{8.9} &\multicolumn{1}{c}{16.08}  &\multicolumn{1}{c}{21.53}\\

& \multicolumn{1}{c}{EC Transferred + NA, $\alpha$ = 5e-8}  &\multicolumn{1}{c}{6.82}  &\multicolumn{1}{c}{8.39} &\multicolumn{1}{c}{16.59}  &\multicolumn{1}{c}{20.88}\\

& \multicolumn{1}{c}{EC Transferred + REG-A}  &\multicolumn{1}{c}{ \ \ \  3.79 $\downarrow$}  &\multicolumn{1}{c}{\ \ \  4.88 $\downarrow$} &\multicolumn{1}{c}{16.13 }  &\multicolumn{1}{c}{ 21.97}\\

& \multicolumn{1}{c}{EC Transferred + REG-B}  &\multicolumn{1}{c}{ \ \ \  4.17 $\downarrow$}  &\multicolumn{1}{c}{\ \ \  5.72 $\downarrow$} &\multicolumn{1}{c}{  16.60 }  &\multicolumn{1}{c}{ 23.19}\\

& \multicolumn{1}{c}{EC Transferred + Adapter}  &\multicolumn{1}{c}{  7.52 }  &\multicolumn{1}{c}{ 9.25 } &\multicolumn{1}{c}{  17.59}  &\multicolumn{1}{c}{   22.85 }\\ 

& \multicolumn{1}{c}{EC Transferred + Adapter + NA, $\alpha$ = 5e-3}  &\multicolumn{1}{c}{8.24}  &\multicolumn{1}{c}{9.79} &\multicolumn{1}{c}{\ \ \  12.76 $\downarrow$}  &\multicolumn{1}{c}{\ \ \  13.15 $\downarrow$}\\

& \multicolumn{1}{c}{EC Transferred + Adapter + NA, $\alpha$ = 5e-4}  &\multicolumn{1}{c}{8.26}  &\multicolumn{1}{c}{10.13} &\multicolumn{1}{c}{20.16}  &\multicolumn{1}{c}{22.51}\\

& \multicolumn{1}{c}{EC Transferred + Adapter + NA, $\alpha$ = 5e-5}  &\multicolumn{1}{c}{7.91}  &\multicolumn{1}{c}{9.64} &\multicolumn{1}{c}{19.37}  &\multicolumn{1}{c}{24.50}\\

& \multicolumn{1}{c}{EC Transferred + Adapter + NA, $\alpha$ = 5e-6}  &\multicolumn{1}{c}{7.44}  &\multicolumn{1}{c}{9.33} &\multicolumn{1}{c}{17.64}  &\multicolumn{1}{c}{22.98}\\

& \multicolumn{1}{c}{EC Transferred + Adapter + REG-A}  &\multicolumn{1}{c}{   8.21}  &\multicolumn{1}{c}{  \ \ \ \ \ \ \ \ \ \ \bf 10.77  \small \small$  \uparrow^{86.3 \%}$} &\multicolumn{1}{c}{  19.93 }  &\multicolumn{1}{c}{  23.99  }\\

& \multicolumn{1}{c}{EC Transferred + Adapter + REG-B}  &\multicolumn{1}{c}{\ \ \ \ \ \ \ \ \ \ \bf 8.44 \small \small$  \uparrow^{97.1 \%}$ }  &\multicolumn{1}{c}{ 10.46} &\multicolumn{1}{c}{ \ \ \ \ \ \ \ \ \ \ \bf  21.59 \small \small$  \uparrow^{41.7 \%}$  }  &\multicolumn{1}{c}{ \ \ \ \ \ \ \ \ \ \ \bf 25.92  \small \small$  \uparrow^{27.3 \%}$}\\ \bottomrule

\end{tabular}
}

\caption{Regularisers with and without annealing.}
\label{table:appendix}
\end{center}
\end{table}

\end{document}